\pdfoutput=1

\documentclass[11pt]{article}

\usepackage[]{ACL2023}

\usepackage{times}
\usepackage{latexsym}
\usepackage{arydshln}

\usepackage[T1]{fontenc}
\usepackage[utf8]{inputenc}

\usepackage{microtype}

\usepackage{inconsolata}
\usepackage{xspace}
\usepackage{amssymb}
\usepackage{multirow}
\usepackage{graphicx}
\usepackage{xcolor}
\usepackage{booktabs}
\usepackage{tabularx}
\usepackage{xr}
\usepackage{color,colortbl}
\usepackage{subcaption}
\definecolor{Gray}{gray}{0.9}
\usepackage{cleveref}
\usepackage{arydshln}

\newcommand{\dataset}{\textsc{Multi}$^3$\textsc{NLU}$^{++}$\xspace}

\newcommand{\hotels}{\textsc{hotels}\xspace}
\newcommand{\banking}{\textsc{banking}\xspace}

\newcommand{\rparagraph}[1]{\vspace{1.3mm}\noindent\textbf{#1.}}
\newcommand{\sparagraph}[1]{\vspace{0.0mm}\noindent\textbf{#1.}}

\newcommand{\sparagraphnodot}[1]{\vspace{0.0mm}\noindent\textbf{#1}}

\newcommand{\tod}{{\textsc{ToD}}\xspace}

\definecolor{Gray}{gray}{0.92}
\newcolumntype{Y}{>{\centering\arraybackslash}X}

\usepackage{todonotes}
\makeatletter
\newcommand*\iftodonotes{\if@todonotes@disabled\expandafter\@secondoftwo\else\expandafter\@firstoftwo\fi}
\makeatother

\definecolor{edolime}{rgb}{0.9,1,0.3}

\title{\dataset: A Multilingual, Multi-Intent, Multi-Domain Dataset for Natural Language Understanding in Task-Oriented Dialogue
}

\author{Nikita Moghe*\textsuperscript{1}\thanks{{ } Equal contribution}, Evgeniia Razumovskaia*\textsuperscript{2}, Liane Guillou\textsuperscript{1}, \\ \bf{Ivan Vulić\textsuperscript{2}, Anna Korhonen\textsuperscript{2},  Alexandra Birch\textsuperscript{1}} \\
  School of Informatics, University of Edinburgh\textsuperscript{1} \\
  Language Technology Lab, University of Cambridge\textsuperscript{2} \\
  {\normalsize \texttt{\{nikita.moghe, liane.guillou, a.birch\}@ed.ac.uk,  \{er563, iv250, alk23\}@cam.ac.uk}}}

\begin{document}
\maketitle

\begingroup\renewcommand\thefootnote{*}
\endgroup

\begin{abstract}
Task-oriented dialogue (\tod) systems have been 
widely deployed in many industries as they deliver more efficient customer support.  These systems are typically constructed for a single domain or language and do not generalise well beyond this.  To support work on Natural Language Understanding (NLU) in \tod across multiple languages \textit{and} domains simultaneously, we constructed \dataset, a \textbf{multi}lingual, \textbf{multi}-intent, \textbf{multi}-domain dataset. \dataset extends the English-only NLU++ dataset to include manual translations into a range of high, medium, and low resource languages (Spanish, Marathi, Turkish and Amharic), in two domains (\textsc{banking} and \textsc{hotels}). 
Because of its multi-intent property, \dataset 
represents complex and natural user goals, and therefore allows us to measure the realistic performance of \tod systems in a varied set of the world's languages. We use \dataset to benchmark state-of-the-art multilingual models 
for the NLU tasks of \textit{intent detection} and \textit{slot labelling} for \tod systems in the multilingual setting. The results demonstrate the challenging nature of the dataset, particularly in the low-resource language setting, offering ample room for future experimentation in multi-domain multilingual \tod setups.
\end{abstract}

\section{Introduction}
Task-oriented dialogue (\tod) systems
\citep{Gupta2006,young-etal-2013-pipeline}, in which conversational agents assist human users to achieve their specific goals, have been used to automate telephone-based and online customer service tasks in a range of domains, including travel \citep{Raux2003, Raux2005}, finance and banking \citep{ALTINOK18.3}, and hotel booking \citep{Li2019}.

ToD systems are often implemented as a pipeline of dedicated modules \citep{Raux2005,young-etal-2013-pipeline}. The \textit{Natural Language Understanding (NLU) module} performs two crucial tasks: \\ 1) intent detection and 2) slot labelling. In the \textit{intent detection} task the aim is to identify or classify the goal of the user's utterance from several \textit{pre-defined classes} (or intents) \citep{tur2010left}. These intents are then used by the \textit
{policy module} \citep{Gašić2012,young-etal-2013-pipeline} to decide the next conversational move by the conversational agent in order to mimic the flow of a human-human dialogue. In the \textit{slot labeling} task each token in an utterance is assigned a label describing the type of semantic information represented by the token. During this process \textit{relevant information} e.g. named entities, times/dates, quantities etc., defining the crucial information of the user's utterance, is identified. 

\begin{figure}[!t]
    \centering
    \includegraphics[width=0.49\textwidth]{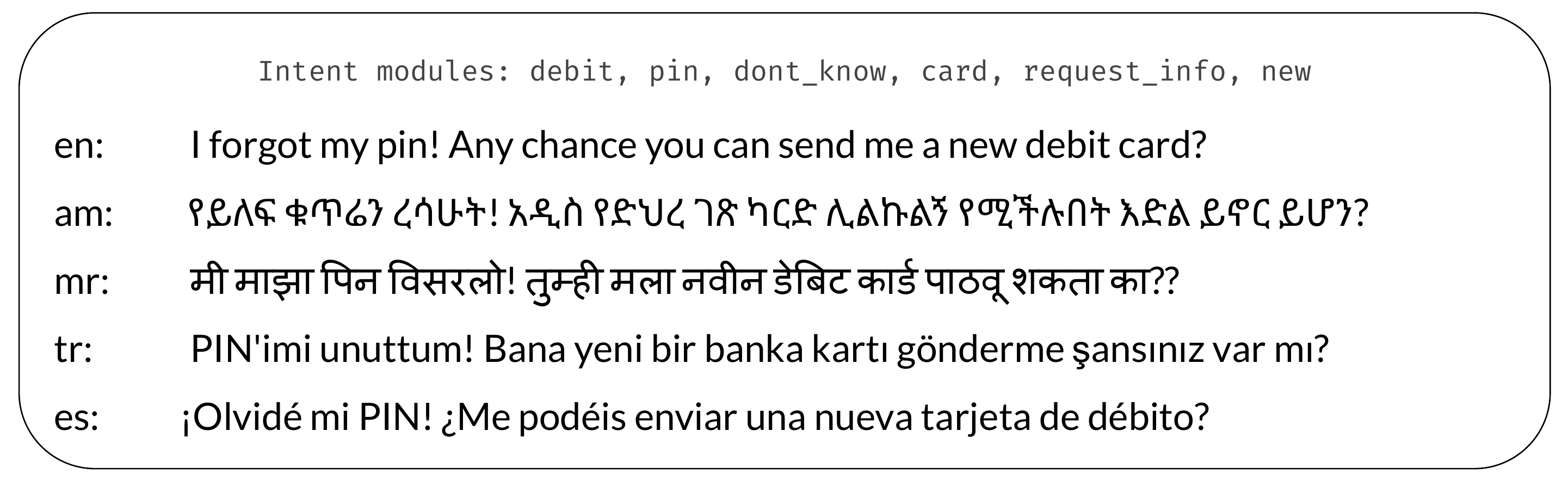}
    \caption{Example from the \dataset dataset for the \banking domain demonstrating the complex NLU tasks of multi-label intent detection across multiple languages. Intent labels consist of generic and domain-specific intents.}
    \label{fig:examples}
    \vspace{-2mm}
\end{figure}

Although intent detection models can reach impressive performance and have been deployed in many commercial systems \cite{ALTINOK18.3,Li2019}, 
they are still unable to fully capture the variety and complexity of natural human interactions, and as such do not meet the requirements for deployment in more complex industry settings \cite{casanueva-etal-2022-nluplusplus}. This is due in part to the limitations of existing datasets for training and evaluating \tod systems. As highlighted by \citet{casanueva-etal-2022-nluplusplus} they are \textbf{1)} predominantly limited to detecting a single intent, \textbf{2)} focused on a single domain, and \textbf{3)} include a small set of slot types \cite{Larson:2022survey}. Furthermore, the success of task-oriented dialogue is \textbf{4)} often evaluated on a small set of higher-resource languages (i.e., typically English) which does not test how generalisable systems are to the diverse range of the world's languages \citep{razumovskaia2022crossing}.

Arguably one of the most serious limitations, hindering their deployment to more complex conversational scenarios, is the inability to handle multiple intents. In many real-world scenarios, a user may express multiple intents in the same utterance, and \tod systems must be able to handle such scenarios \cite{gangadharaiah-narayanaswamy-2019-joint}. For example, in Figure~\ref{fig:examples}, the user expresses two main intents: (i) informing that they have forgotten their pin and thus, (ii) they would like to request a new debit card instead. A single-intent detection system can detect either of the two intents (but not both), resulting in partial completion of the user's request. \citet{casanueva-etal-2022-nluplusplus} recently proposed a multi-label intent detection dataset to capture such complex user requests. They further propose using intent modules as intent labels that can act as sub-intent annotations. In this example, ``pin'' and ``don't\_know'' compose the first intent while  ``request\_info'', ``new'', ``debit'', and ``card''  compose the second. The use of intent modules, due to their combinatorial power, can support more complex conversational scenarios, and also allows reusability of the annotations across multiple domains. For example, ``request\_info'', ``new'', and ``membership'' can be reused for gyms, salons, \textit{etc.} to request information about new memberships at the respective institutions.

Furthermore, ToD systems are typically constructed for a single language. Their extension to other languages is restricted by the lack of available training data for many of the world's languages. 
Whilst the construction of multilingual \tod 
datasets has been given some attention \citep{razumovskaia2022crossing,majewska2022cross,xu2020end}, these datasets often include synthetic translations in the form of post-edited Machine Translation output \cite{ding-etal-2022-globalwoz,zuo2021allwoz,Bellomaria2019}. Post-editing may introduce undesirable effects and result in texts that are simplified, normalised, or exhibit interference from the source language as compared with manually translated texts \citep{toral-2019-post}.

To address all of the limitations discussed above, we propose \dataset, a \textbf{multi}lingual, \textbf{multi}-intent, \textbf{multi}-domain for training and evaluating \tod systems. \dataset extends the recent monolingual English-only dataset NLU++, which is a multi-intent, multi-domain dataset for the \banking and \hotels domains. \dataset adds the element of multilinguality and thus enables simultaneous cross-domain and cross-lingual training and experimentation for \tod NLU as its unique property.

\dataset includes expert manual translations of the 3,080 utterances in NLU++ to four languages of diverse typology and data availability: Spanish, Marathi, Turkish, and Amharic. The selection of languages covers a range of language families and scripts and includes high, medium, and low-resource languages. Capturing language diversity is particularly important if we wish to design multilingual \tod systems that are robust to the variety of expressions used across languages to represent the same value or concept. 
Using \dataset we demonstrate the challenges involved in extending existing state-of-the-art Machine Translation systems and multilingual language models for NLU in \tod systems. \dataset is publicly available at \url{https://huggingface.co/datasets/uoe-nlp/multi3-nlu}.

\section{Background and Related Work}

\sparagraphnodot{NLU++}
\citep{casanueva-etal-2022-nluplusplus}, which serves as the base for \dataset, covers two domains: \banking and \hotels. The intent and slot ontologies include intents and slots which are general, cross-domain types, as well as domain-specific ones. The intent ontology includes  a total of 62 intents of which 23 and 14 intents are \banking and \hotels specific, respectively. The slot ontology includes 17 slot types, of which three and four slot types are \banking and \textsc{hotels} specific, respectively. Such intent and slot ontology construction allows for the extension to new domains more easily. In this work, we inherit monolingual NLU++'s core benefits over previous datasets with the additional layer of multilinguality: \textbf{1)} it is based on real customer data which addresses the issue of low lexical diversity in crowd-sourced data \citep{Larson:2022survey}, \textbf{2)} it supports the requirement for production systems to capture multiple intents from a single utterance, and \textbf{3)} it is also \textit{slot-rich} -- it combines a large set of fine-grained intents with a large set of fine-grained slots to facilitate more insightful evaluation of models that jointly perform the two NLU tasks \citep{Chen2019Joint,gangadharaiah-narayanaswamy-2019-joint}.

\rparagraph{Multilingual NLU Datasets}
Prior work has demonstrated the importance and particular challenges posed by low-resource languages \citep{goyal-etal-2022-flores, magueresse2020low, xia2021metaxl}, while an increasing number of multilingually pretrained models \cite{xue2021mt5, conneau-etal-2020-unsupervised, feng-etal-2022-language, liu2020multilingual} enable significant improvements in processing them. While some tasks (e.g., NER \citep{adelani2021masakhaner} or NLI \citep{ebrahimi-etal-2022-americasnli}) already have benchmarks to evaluate on low-resource languages, dialogue Natural Language Understanding (NLU) is still lagging behind in this respect. The reasons for this include the high cost of data collection, as well as specific challenges posed by dialogue, e.g., colloquial speech or tone used in live conversations. Additionally, while we have observed increased interest in few-shot methods for multilingual dialogue NLU \cite[\textit{among others}]{bhathiya2020meta, moghe-etal-2021-cross, feng-etal-2022-language, razumovskaia2022data},  none of the existing datasets \cite{xu2020end, fitzgerald2022massive, van2021masked} allow for reproducible comparison between few-shot methods. In other words, no dataset to date has provided predefined splits for few-shot experiments.
\begin{table*}[!t]
\small
\centering
\begin{tabular}{@{}lcccccc@{}}
\toprule
Dataset                                                         & \# Languages & Domains & Size                                                                                   & \# intents & \# slots & Multi-Intent? \\ \midrule
\begin{tabular}[c]{@{}l@{}}Multilingual TOP\\ \citep{schuster2019cross}\end{tabular} & 3            & 3       & \begin{tabular}[c]{@{}l@{}}43,323 {[}en{]}\\ 8,643 {[}es{]}\\ 5,082 {[}th{]}\end{tabular} & 12         & 11     & $\times$  \\ \cmidrule(lr){1-7}
\begin{tabular}[c]{@{}l@{}}MultiATIS++\\ \citep{xu2020end}\end{tabular}      & 8            & 1       & 5,871                                                                                   & 18         & 84    & $\times$   \\ \cmidrule(lr){1-7}
\begin{tabular}[c]{@{}l@{}}xSID\\ \citep{van2021masked}\end{tabular}             & 13           & 6       & 800                                                                                    & 16         & 33   & $\times$    \\ \cmidrule(lr){1-7}
\begin{tabular}[c]{@{}l@{}}MASSIVE\\ \citep{fitzgerald2022massive}\end{tabular}          & 51           & 18      & 19,521                                                                                  & 60         & 55  & $\times$     \\ \cmidrule(lr){1-7}
\dataset                                                     & 4            & 2       & 3,080                                                                                   & 62         & 17    & \checkmark   \\ \bottomrule
\end{tabular}%
\caption{Statistics of representative multilingual dialogue NLU datasets. MultiATIS++ contains some utterances with more than one intent label but the vast majority are single-label.}\label{tab: nlu multilingual datasets}
\vspace{-1.5mm}
\end{table*}

Until recently, resources for multilingual NLU have been scarce \citep{razumovskaia2022crossing}. The vast majority were built upon the English ATIS dataset \citep{price1990atis}, which has been extended to ten target languages \citep{upadhyay2018almost, dao2021intent, xu2020end}. However, ATIS covers only one domain (airline booking) and has been claimed to be almost solved already, more than a decade ago \citep{tur2010left}. More recent datasets cover multiple domains and broader linguistic geography \cite{majewska2022cross,schuster2019cross,fitzgerald2022massive}, including domains such as music or alarm, which are in frequent use in production systems.

\textit{All} of the existing multilingual NLU datasets label every user utterance with a single intent, although current production-level systems often rely on multi-intent labelling \cite{gangadharaiah-narayanaswamy-2019-joint,qin2020agif}, allowing for faster development cycles and updates if a new, previously unseen intent is observed \citep{casanueva-etal-2022-nluplusplus}. \dataset is the first multilingual, multi-intent dataset with modular intent annotations. In comparison to other multilingual NLU datasets presented in Table~\ref{tab: nlu multilingual datasets}, \dataset has a larger intent set and is natively multi-intent/label. Unlike xSID, which is an evaluation-only dataset, it contains both training and evaluation data for all languages. Further, while MASSIVE is based on utterances specifically generated for the dataset \citep{bastianelli2020slurp}, \dataset is based on \textit{real user inputs} to a system in industrial settings \cite{casanueva-etal-2022-nluplusplus}. 
\dataset enables systematic comparisons of dialogue NLU systems in few-shot setups for cross-lingual and cross-domain transfer for low-, medium- and high-resource languages.


\section{Dataset Collection}

Our data collection process focuses on creating datasets with natural and realistic conversations, avoiding many artefacts that arise from crowdsourcing or automatic translation. We ask professional translators to manually translate each source-language utterance into the four target languages in the dataset; this promotes equal opportunity for future research for all four languages, and enables comparative cross-language analysis. 


\rparagraph{Choice of Base Dataset} This work aims to create a multilingual dataset which would be useful for testing production-like systems in multiple languages.  Thus, we chose NLU++ \citep{casanueva-etal-2022-nluplusplus} as our base dataset because i) it consists of real-world examples; and ii) every example is labelled with multiple intents which has proven useful in production settings \cite{qin2020agif, casanueva-etal-2022-nluplusplus}. 
  
Prior work on English multi-intent classification often uses MixSNIPS and MixATIS as benchmarks \citep{qin2020agif}. However, the multi-intent examples in these datasets are synthetic, i.e., they are obtained through a simple concatenation of two single-intent examples, e.g., ``Play this song and book a restaurant''. This process leads to repetitive content and unnatural examples in the datasets. In contrast, NLU++ consists of intrinsically multi-intent examples such as ``I cannot login to my account because I don't remember my pin.'', and is much more semantically variable.

\rparagraph{Languages}
We selected Spanish, a widely-spoken Romance language, as our high-resource language. Marathi, an Indo-Aryan language predominantly spoken in the Indian state of Maharashtra, is our medium-resource language. Turkish, an agglutinative Turkic language, may be regarded as a low-resource language from a Machine Translation perspective, or as a medium-resource language based on the amount of training data in XLM-R \cite{conneau-etal-2020-unsupervised}.\footnote{The amount of Turkish data is higher than, e.g., Marathi (which we consider to be a medium-resource language).} Amharic, an Ethiopian Semitic language belonging to the Afro-Asiatic language family, is our low-resource language. Spanish and Turkish are written in Latin script, Marathi is written in Devanagari, and Amharic in Ge'ez script. 


\rparagraph{Manual Translation}
The use of crowd-sourcing agencies to collect multilingual datasets has often resulted in crowd workers using a Machine Translation (MT) API (plus post-editing) to complete the task, or even simply transliterating the given sentence \citep{goyal-etal-2022-flores}. In the case of post-editing MT output, translations may exhibit \textit{post-editese} -- post-edited sentences are often simplified, normalised, or exhibit a higher degree of interference from the source language than manual translations \citep{toral-2019-post}. In our case, we wish to preserve the register of the original utterances, in particular with respect to the colloquial nature of many of the utterances in the original English NLU++ dataset.
Furthermore, we wish to collect high-quality, natural translations. We, therefore, opted to recruit professional translators to perform manual translation, via two online platforms: Proz.com\footnote{\url{www.proz.com/};  Proz.com is a platform for recruiting freelance translators who self-quote their remuneration.} and Blend Express.\footnote{\url{www.getblend.com/online-translation/}} 
We instructed our translators to treat the task as \textit{a creative writing task} \cite{ponti-etal-2020-xcopa} and maintain the \textit{colloquial nature of the utterances}. We also provided instructions to annotate the spans in the generated translations. We provide the instructions given to our translators in Appendix~\ref{app:translation_guidelines}. We recruited three translators per language; Spanish translators were recruited via Proz.com, and translators for the remaining languages were assigned to us by Blend Express.

We first conducted a pilot task in which we asked professional translators to translate and annotate 50 sentences per domain. We conducted an in-house evaluation by native speakers of the respective language to verify that the translations were colloquial, that named entities were appropriately translated, and that the translation was of high quality. These evaluations were communicated to the translators.\footnote{As translators are generally not used to annotating spans in text, a significant amount of time was spent on training the translators for this new task.}

After the pilot, we asked the same translators to complete the translation of the remaining utterances in the dataset. We ran an automatic checker to ensure that the slot values marked by the translators were present within their translated sentences. Further corrections such as incorrect annotations were also communicated to the translators.



\rparagraph{Duration and Cost}
Data collection was carried out over five months and involved (i) selecting translation agencies, (ii) running the pilot task, (iii) providing feedback to the translators, and (iv) the full-fledged data collection phase.
The professional translators were compensated at £0.06/word for Spanish and £0.07/word for the remaining languages. The total cost of \dataset is £7,624; see Appendix~\ref{sec:annotation_costs} for further details.

\rparagraph{Slot Span Verification} Initially, the translators performed slot labelling simultaneously with translation. To ensure the quality of the annotation, the translated data was revised and annotated by three native speakers of the target language. Similar to \citet{majewska2022cross}, we used an automated inter-annotator reliability method to automatically verify the annotation quality. We conducted slot span labelling for 200 examples in Spanish for both \textsc{Banking} and \textsc{hotels} domains from three native Spanish annotators. The accuracy score\footnote{The accuracy is the ratio between the number of exactly matching spans between the annotators to the total number of slot values annotated.} for a given sub-sample was at 86.8\% revealing high agreement between the annotators. 
\section{Baseline Experiments}

We benchmark several state-of-the-art approaches on \dataset to \textbf{1)} provide reference points for future dataset use and \textbf{2)} demonstrate various aspects of multilingual dialogue NLU systems which can be evaluated using the dataset, such as cross-lingual and cross-domain generalisation. We provide baseline numbers for intent detection and slot labelling and
analyse the performance across languages and methods, and for different sizes of training data. We follow the main experimental setups from \citet{casanueva-etal-2022-nluplusplus} where possible, but we extend them to multilingual contexts.

\rparagraph{Training Data Setups} We follow an N-fold cross-validation setup following the setup in \citet{casanueva-etal-2022-nluplusplus}.
The experiments were run in three setups: \textit{low}, \textit{mid} and \textit{large}. The \textit{low} data setup corresponds to 20-fold cross-validation, where the model is trained on  $\frac{1}{20}$th of the dataset and tested on the remaining 19 folds. The \textit{mid} and \textit{large} setups correspond to 10-fold cross-validation, where in \textit{mid} setup the model is trained on $\frac{1}{10}$th of the data and tested on the other nine folds, and vice versa for the \textit{large} setup. 

\rparagraph{Domain Setups} \dataset contains training and evaluation data for two domains: \banking and \hotels. It thus enables evaluation of NLU systems in three domain setups: \textit{in-}domain, \textit{cross-}domain and \textit{all-}domain. In the \textit{in-}domain setup, the model is trained and evaluated on the same domain, that is, we are testing how well the model can generalise to unseen user utterances while operating in the same intent space as the training data and without any domain distribution shift. In the \textit{cross-}domain setup, the model is trained on one domain and tested on the other domain. We evaluate on the union of label sets of two domains rather than on the intersection as done by \citet{casanueva-etal-2022-nluplusplus}. In this setup, we are testing how well a model can generalise to a new, unseen domain including intents unseen in training. In the \textit{all-}domain setup, we train and evaluate the models on data from both domains. In this setup, we are testing how models perform on the larger label set (where some labels are shared between the domains) when examples are provided for all classes.

\rparagraph{Language Setups} 
\dataset, offering comparable sets of annotated data points across languages, allows for systematic comparisons of multilingual dialogue NLU systems on languages with different amounts of resources and diverse typological properties. We evaluate NLU systems in two setups: \textit{in-}language and \textit{cross-}lingually. \textit{Cross-}lingual benchmarking is conducted with the established approaches: (i) direct transfer using multilingually pretrained large language models (e.g.,  XLM-R; \citet{conneau-etal-2020-unsupervised}), and (ii) transfer via translation, i.e., when either the test utterances are translated into the source language (\textit{Translate-Test}). 
We source our translations from the M2M100 translation model \citep{fan-etal-2021-m2m100}.

\begin{table*}[!t]\centering
\small
\def\arraystretch{0.86}
\begin{tabularx}{\linewidth}{l YY YY YY YY YY YY}\toprule
&\multicolumn{2}{c}{\textbf{\textsc{english}}} &\multicolumn{2}{c}{\textbf{\textsc{amharic}}} &\multicolumn{2}{c}{\textbf{\textsc{marathi}}} &\multicolumn{2}{c}{\textbf{\textsc{spanish}}} &\multicolumn{2}{c}{\textbf{\textsc{turkish}}} &\multicolumn{2}{c}{\textbf{\textsc{avg}}} \\\cmidrule(lr){2-3} \cmidrule(lr){4-5} \cmidrule(lr){6-7} \cmidrule(lr){8-9} \cmidrule(lr){10-11} \cmidrule(lr){12-13}
&\textsc{H} & \textsc{B} &\textsc{H} & \textsc{B} &\textsc{H} & \textsc{B} &\textsc{H} &\textsc{B} &\textsc{H} & \textsc{B}  & \textsc{H} & \textsc{B} \\\midrule
Full-Fine-tuning: XLM-R &50.0 &61.7 &43.7 &51.3 &46.4 &57.9 &48.3 &59.3 &49.9 &56.9 &47.6 &55.8 \\
MLP-based: LaBSE &55.4 &64.3 & \textbf{48.8} & \textbf{56.6} & \textbf{53.5} & \textbf{63.5} &56.6 &64.0 &55.9 & \textbf{61.6} &54.1 & \textbf{60.7} \\
MLP-based: mpnet &\textbf{61.8} & \textbf{68.8} &39.9 &44.1 &53.1 &58.8 & \textbf{61.1} & \textbf{65.2} & \textbf{59.6} &60.3 & \textbf{55.1} &58.7 \\
\bottomrule
\end{tabularx}
\vspace{-1mm}
\caption{Comparison between \textit{Full fine-tuning} and \textit{MLP-based} setups for intent detection ($F_1$ $\times$ 100). The experiments are run for the \textbf{10-Fold} in-language in-domain setup. \textsc{H} is \hotels domain and \textsc{B} is \banking domain.}\label{tab: comparison finetuning mlp based}
\vspace{-1.5mm}
\end{table*}

\begin{table*}[!t]\centering
\small
\def\arraystretch{0.9}
\begin{tabularx}{0.95\linewidth}{l YYY YYY YYY}
\toprule
  {} & \multicolumn{3}{c}{\bf \textsc{banking}} & \multicolumn{3}{c}{\bf \textsc{hotels}} & \multicolumn{3}{c}{\bf \textsc{all}} \\
  \cmidrule(lr){2-4} \cmidrule(lr){5-7} \cmidrule(lr){8-10}
\textbf{Model} & \textbf{20-Fold}  & \textbf{10-Fold} & \textbf{Large} & \textbf{20-Fold}  & \textbf{10-Fold} & \textbf{Large} & \textbf{20-Fold}  & \textbf{10-Fold} & \textbf{Large} \\ \midrule
\rowcolor{Gray} {} & \multicolumn{9}{c}{\bf \textsc{amharic}} \\ \midrule
LaBSE & \textbf{46.2} & \textbf{56.6} & \textbf{77.3} & \textbf{38.6} & \textbf{48.8} & \textbf{71.9} & \textbf{46.4} & \textbf{56.1} & \textbf{76.1} \\
mpnet &37.0 &44.1 &64.7 &31.9 &39.9 &60.6 &36.2 &43.1 &63.5 \\ \midrule
\rowcolor{Gray} {} & \multicolumn{9}{c}{\bf \textsc{spanish}} \\ \midrule
LaBSE &51.4 &64.0 & \textbf{85.2} &46.4 &56.6 &79.5 &52.2 &63.5 & \textbf{83.8} \\
mpnet & \textbf{55.2} & \textbf{65.2} &84.4 & \textbf{50.9} & \textbf{61.1} & \textbf{80.3} & \textbf{55.4} & \textbf{65.1} &83.5 \\
\bottomrule
\end{tabularx}
\vspace{-1mm}
\caption{In-language in-domain intent detection performance for Amharic and Spanish  ($F_1$ $\times$ 100). Results for other languages are provided in Appendix~\ref{app: mlp results}, Table~\ref{tab: in-domain in-language}.}
\label{tab: main in-domain in-language}
\vspace{-1.5mm}
\end{table*}

\subsection{Classification-Based Methods} 
\label{sec:MLP-intent}
\rparagraph{Experimental Setup and Hyperparameters} We evaluate two standard classification approaches to intent detection: (i) MLP-based with a fixed encoder; and (ii) full-model fine-tuning.
Prior work has demonstrated that strong intent detection results can be attained without fine-tuning the full encoder model both in monolingual \cite{casanueva2020efficient} and multilingual setups \cite{gerz2021multilingual}. The idea is to use a fixed efficient sentence encoder to encode sentences and train only the multi-layer perceptron (MLP) classifier on top of that to identify the intents. As we are dealing with multi-label classification, a sigmoid layer is stacked on top of the classifier. Intent classes for which the probability scores are higher than a (predefined) threshold are considered active. As in \citep{casanueva-etal-2022-nluplusplus}, we use the threshold of 0.3. In the experiments we evaluate two state-of-the-art multilingual sentence encoders: 1) \textit{mpnet}, a multilingual sentence encoder trained using multilingual knowledge distillation (\citet{reimers2020making}), and 2) \textit{LaBSE}, a language-agnostic BERT sentence encoder \citep{feng-etal-2022-language} which was trained using dual-encoder training. LaBSE was especially tailored to produce improved sentence encodings in low-resource languages. The models were loaded from the \texttt{sentence-transformers} library \cite{reimers-2019-sentence-bert}. 

In the full fine-tuning setup, which is the current standard in multilingual NLP \cite{xu2020end}, a multilingually pretrained encoder is fine-tuned on task-specific data. In our case, XLM-R (Base) \cite{conneau-etal-2020-unsupervised} is fine-tuned for multi-label intent detection. 

All MLP-based models were trained with the same hyperparameters, following the suggested values from \citet{casanueva-etal-2022-nluplusplus}. The MLP classifier comprises one 512-dimensional hidden layer and $tanh$ as non-linear activation. The learning rate is 0.003 for MLP-based models and $2$e-$5$ for full model fine-tuning, with linear weight decay in both setups. For all setups, the models were trained with AdamW~\cite{adamw} for 500 and 100 epochs for intent detection and slot labelling, respectively. We used a batch size of 32. The evaluation metric is micro-$F_1$.

\begin{table*}[!t]
\def\arraystretch{0.83}
\centering
{\small
\begin{tabularx}{0.95\linewidth}{ll Y|YYYY ll  Y|YYYY}
\toprule
  {} & {} & \multicolumn{5}{c}{\bf \textsc{TGT}} & {} & {} & \multicolumn{5}{c}{\bf \textsc{TGT}} \\
  \cmidrule(lr){3-7} \cmidrule(lr){10-14} 
\textbf{SRC} & \textbf{DOMAIN}& \textsc{en} & \textsc{am}  & \textsc{mr} & \textsc{es} & \textsc{tr}  & \textbf{SRC} & \textbf{DOMAIN} &  \textsc{am} & \textsc{en} & \textsc{mr} & \textsc{es}  &  \textsc{tr} \\
\cmidrule(lr){1-7}\cmidrule(lr){8-14}

\multirow{3}{*}{\textsc{en}} & \textsc{banking} & 51.6 & 38.9 & 34.1 & 36.2 & 29.5 & \multirow{3}{*}{\textsc{am}} & \textsc{banking} &46.2  &48.4 &38.9 &40.2 & 33.8 \\
 & \textsc{hotels} & 45.1 & 29.8 &29.2 &34.4 & 27.5 & 
 & \textsc{hotels} & 38.6 &43.5 &33.5 &37.4 &31.3 \\
 & \textsc{all} & 51.6 &37.7 &32.9 &35.9 & 28.8 & 
 & \textsc{all} & 46.4 &48.4 &39.1 &41.5 & 34.6 \\
\bottomrule
\end{tabularx}
}%
\vspace{-1.5mm}
\caption{Cross-lingual in-domain results for intent detection with MLP-based and LaBSE as the fixed sentence encoder ($F_1$ $\times$ 100). The results are presented for transfer from English and Amharic for the \textbf{20-Fold} setup, while the results for other source languages and training data setups are available in Appendix~\ref{app: mlp results}, Table~\ref{tab: cross-domain in-language}. }\label{tab: main in-domain cross-language}
\vspace{-1.5mm}
\end{table*}

\begin{table*}[!t]
\def\arraystretch{0.83}
\centering
{\small
\begin{tabularx}{0.95\linewidth}{ll Y|YYYY ll  Y|YYYY}
\toprule
  {} & {} & \multicolumn{5}{c}{\bf \textsc{TGT}} & {} & {} & \multicolumn{5}{c}{\bf \textsc{TGT}} \\
  \cmidrule(lr){3-7} \cmidrule(lr){10-14} 
\textbf{SRC} & \textbf{DOMAIN}& \textsc{en} & \textsc{am}  & \textsc{mr} & \textsc{es} & \textsc{tr}  & \textbf{SRC} & \textbf{DOMAIN} &  \textsc{am} & \textsc{en} & \textsc{mr} & \textsc{es}  &  \textsc{tr} \\
\cmidrule(lr){1-7}\cmidrule(lr){8-14}

\multirow{3}{*}{\textsc{en}} & \textsc{banking} & 37.3 & 15.3 & 19.7 & 22.9 & 18.8 & \multirow{3}{*}{\textsc{am}} & \textsc{banking} & 27.3 & 19.9 & 22.8 & 15.5 & 20.4 \\
 & \textsc{hotels} & 24.3 & 6.1 &10.8 &14.3 & 14.3 & 
 & \textsc{hotels} &10.4 &  7.9 & 6.4 & 5.6 & 6.8 \\
 & \textsc{all} & 43.7 & 16.4 & 22.0 & 28.4 & 26.3 & 
 & \textsc{all} & 28.6 & 20.0 & 21.9 & 15.4 & 21.6 \\
\bottomrule
\end{tabularx}
}%
\vspace{-1.5mm}
\caption{Cross-lingual in-domain results for slot labelling with full fine-tuning of XLM-R ($F_1$ $\times$ 100). The results are presented for transfer from English and Amharic for the \textbf{20-Fold} setup, while the results for other source languages and training data setups are available in Appendix~\ref{app: slot labeling results}, Table~\ref{tab: token classification  cross-lingual slot labelling}. }
\label{tab: main in-domain cross-language slot labelling}
\vspace{-1.5mm}
\end{table*}

\rparagraph{Results and Discussion} Table~\ref{tab: comparison finetuning mlp based} presents the comparison between the \textit{full fine-tuning} based on XLM-R and \textit{MLP-based} classification approaches. The results demonstrate that for the in-domain in-language 10-fold setup, the \textit{MLP-based} approach works consistently better than \textit{full fine-tuning} across domains and languages. We assume that the reason is that the \textit{MLP-based} approach is more parameter-efficient than \textit{full fine-tuning}, making it more suited for such low-data setups. Due to their computational efficiency combined with competitive performance, we focus on MLP-based models for the remainder of this section. 

The main MLP-based intent detection results are presented in Tables~\ref{tab: main in-domain in-language} and~\ref{tab: main in-domain cross-language} for in-domain {intent detection} for the in-language and zero-shot cross-lingual setups, respectively. When we compare the performance on low- and high-resource languages, although the sizes and content of the training data are the same across languages, we observe a large gap between the performance of the same models on Spanish and Amharic. In addition, the results in Table~\ref{tab: main in-domain in-language} reveal the properties of the multilingual sentence encoders with respect to the intent detection task. While \textit{mpnet} performs consistently better on Spanish (our high-resource language), \textit{LaBSE} is a much stronger encoder for low-resource Amharic. The differences are especially pronounced in the low-data setups (\textbf{20-Fold} and \textbf{10-Fold}). While for Spanish this difference can be recovered with more training data (cf. \textbf{Large} setup), for Amharic these differences persist and are amplified across large training data setups.

We consider zero-shot transfer from two source languages: English (the most commonly used high-resource source language) and Amharic (our low-resource option). Surprisingly, the results in Table~\ref{tab: main in-domain cross-language} suggest that using English as a default source language, as typically done in work on cross-lingual transfer, might not be optimal. In fact, using Amharic as a source language leads to stronger transfer results across languages. One trend we observed is that the lower resource the source language is, the stronger the performance on diverse target languages.\footnote{This trend is observed across other source languages and both sentence encoders, as shown in Appendix~\ref{app: mlp results}.} We speculate that by training on lower-resource language data we might `unearth' the sentence encoders' multi-lingual capabilities.

The main results for {slot labelling} are presented in Table~\ref{tab: main in-domain cross-language slot labelling} for the in-domain in-language and zero-shot cross-lingual setups. The comparison of in-language results for high-resource (en) and low-resource (am) demonstrate a similar trend to intent classification: in-language performance on the  high-resource language is stronger than that on the low-resource language. However, unlike for intent detection, the high-resource language serves as a stronger source for cross-lingual transfer than the low-resource language.

\subsection{QA-Based Methods}

\sparagraph{Experimental Setup and Hyperparameters} 
Reformulating \tod tasks as question-answering (QA) problems has achieved state-of-the-art performance \citep{DBLP:conf/icassp/NamazifarPTH21}. As these methods were the best performing for the NLU++ dataset, we now investigate this approach in the multilingual setting for both NLU tasks. 
To formulate intent detection as an extractive QA task, the utterance is appended with ``yes. no. [UTTERANCE]'' and acts as the context. 
Intent labels are converted into questions as ``Is the intent to ask about [INTENT\_DESCRIPTION]'' where INTENT\_DESCRIPTION is the free-form description of the intent. 
The QA model must learn to predict the span as ``yes'' across all the questions corresponding to the specific intents in the utterance and ``no'' otherwise. During the evaluation of transfer performance to other languages, the utterance is in the target language while the question is in the source language.~\footnote{We found this setup to be empirically better than posing the question and utterance in the same language.}
For slot labelling, the template is ``none. [UTTERANCE]''. We follow the strict evaluation approach for slot labelling - a span is marked as correct only if it exactly matches the ground truth span \citep{DBLP:conf/icassp/NamazifarPTH21}. 

To build an extractive QA model for these tasks, we first fine-tune a multilingual language model mDeBERTa \citep{deberta} with a general-purpose QA dataset such as SQuADv2 \citep{rajpurkar-etal-2018-know} and then with the respective languages from our dataset. We fine-tune for 10 epochs (5 for the Large setting) with a learning rate of $1e-5$, weight decay is 0, and batch size is 4.  
\begin{table*}[!t]
\centering
\small
\def\arraystretch{0.8}
\begin{tabular}{@{}llrrrrr@{}}
\toprule
\def\arraystretch{0.8}
\textbf{Method} & \textbf{Domain} & \multicolumn{1}{l}{\textsc{en}} & \multicolumn{1}{l}{\textsc{\sc am}} & \multicolumn{1}{l}{\sc mr} & \multicolumn{1}{l}{\sc es} & \multicolumn{1}{l}{\sc tr} \\ \midrule
\multicolumn{7}{c}{\bf Cross-lingual zero-shot transfer} \\ \midrule

\multirow{3}{*}{\sc en} & \banking & 80.8  $\pm$ 1.3 & 42.7 $\pm$ 4.5& 58.4 $\pm$ 3.2 & 66.2 $\pm$ 3.1 & 56.5 $\pm$ 3.6 \\
 & \hotels & 66.9 $\pm$ 3.0 & 37.1 $\pm$ 3.3 & 51.1 $\pm$ 2.9 & 62.2 $\pm$ 2.5 & 59.6 $\pm$ 2.8 \\
 & \textsc{all} & 79.6 $\pm$ 1.3 & 40.7 $\pm$ 4.8 & 56.5 $\pm$ 3.7 & 67.9 $\pm$ 3.5 & 58.2 $\pm$ 4.3 \\ \midrule
\multirow{3}{*}{\sc am} & \banking  & 56.1 $\pm$ 2.2 & 30.1 $\pm$ 4.7 & 31.3 $\pm$ 4.5 & 27.9  $\pm$ 4.9 & 28.5 $\pm$ 4.7 \\
 & \hotels & 30.5 $\pm$ 8.4 & 51.1 $\pm$ 2.1& 27.5 $\pm$ 7.6 & 31.5 $\pm$ 8.3 & 33.0 $\pm$ 8.5 \\
 & \textsc{all} & 29.4 $\pm$ 4.9 & 57.2 $\pm$ 2.6 & 32.3 $\pm$ 4.1 & 27.7 $\pm$ 5.2  & 30.7  $\pm$ 4.5\\ \midrule
\multicolumn{7}{c}{\bf Translate-Test} \\ \midrule
\textsc{en} & \textsc{all} & - & 16.2 $\pm$ 3.0 & 63.3 $\pm$ 2.1 & 67.8 $\pm$ 1.8 & 46.3 $\pm$ 5.0\\ 
\textsc{am} & \textsc{all} & 16.2 $\pm$ 4.1 & - & 15.2 $\pm$ 4.0 & 11.7 $\pm$ 2.5 & 11.7 $\pm$ 2.5\\ \midrule

\end{tabular}
\vspace{-1mm}
\caption{$F_1$ $\times$ 100 for the QA-based intent detection models in the \textbf{20-Fold} setup using \textsc{en} and \textsc{am} for training. The remaining results are  in Appendix \ref{app: qa intent results}.}
\label{tab:qa_intents_main}
\vspace{-2.5mm}
\end{table*}


\begin{table}[!t]
\centering
\def\arraystretch{0.8}
\small
\begin{tabular}{@{}lr|rrrr@{}}
\toprule
\textbf{Domain} & \multicolumn{1}{l}{\textsc{en}} & \multicolumn{1}{l}{\textsc{am}} & \multicolumn{1}{l}{\textsc{mr}} & \multicolumn{1}{l}{\textsc{es}} & \multicolumn{1}{l}{\textsc{tr}} \\ \midrule
\textsc{banking} & 59.7 & 44.2 & 48.7 & 42.9 & 52.3 \\
\textsc{hotels} & 61.6 & 47.6 & 42.0 & 49.0 & 50.9 \\
\textsc{all} & 63.0 & 46.8 & 46.9 & 47.7 & 53.6 \\ \bottomrule
\end{tabular}
\vspace{-0.5mm}
\caption{$F_1$ $\times$ 100 for cross-lingual zero-shot results of QA-based slot labelling models when trained with English data for the \textbf{20-Fold} setup. The results for other languages and data setups are in Appendix \ref{app: slot labeling results}.}
\label{tab:value-qa}
\vspace{-1.5mm}
\end{table}
\rparagraph{Results and Discussion} 
We report the results for intent detection in the 20-Fold setup using English and Amharic in Table~\ref{tab:qa_intents_main}, and the remaining results are in Appendix \ref{app: qa intent results}. We also compare cross-lingual transfer with translation-based methods. We find that mDeBERTa-based QA models have comparable results for English multi-intent detection even with the monolingual models in \citet{casanueva-etal-2022-nluplusplus}. We notice a drastic drop in the zero-shot transfer performance across all languages and domains. This is in line with recent findings that zero-shot transfer is harder for dialogue tasks \citep{ding-etal-2022-globalwoz, hung-etal-2022-multi2woz, majewska2022cross} as opposed to other cross-lingual tasks \citep{hu-etal-xtreme}. Unlike the observation in \S\ref{sec:MLP-intent}, using 
a higher resource language is  beneficial for transfer learning in the QA setup. 
We find that improvement in transfer performance depends more on the matching of the script of the source and target language followed by the amount of training data present per language during the pre-training of the mDeBERTa model. The exception is Amharic as the source language, where transfer performance is poor across the board. The scores also indicate QA models are better at this task than MLP methods, cf. Table \ref{tab: main in-domain cross-language slot labelling}. 

While comparing zero-shot cross-lingual transfer with Translate-Test, we find that the latter is poorer (except in the case of \textsc{en}-\textsc{mr}), opposing the findings in \citet{majewska2022cross}. We attribute this to the poor quality of the translations as repetition/nonsensical generation is rampant in them. Further, translation-based methods have an additional overhead of translating the sentences.  

We find that the standard deviation is quite variable across both tasks. We check for standard deviations across different data setups and find that the trend is consistent irrespective of the underlying training data. There is also no visible trend between the language similarity or the amount of training data present during pre-training of mDeBERTa.

Table \ref{tab:value-qa} reports the slot labelling results. Unlike intent detection results, slot labelling results are comparable across the languages in the same data settings. Similar to the intent detection results, using mDeBERTa QA models yields better results than those reported in Table~\ref{app: slot labeling results} for high resource languages. Overall, we find that QA-based methods are indeed a promising research avenue \citep{DBLP:journals/corr/abs-2107-13586}, even in the multilingual setting~\citep{zhao-schutze-2021-discrete}.\footnote{The efficiency overhead of the QA-based approach is that every utterance is accompanied by $X$ questions where $X$ is the number of intent/slot labels.} 


\section{Discussion and Future Work}
\captionsetup[subfigure]{oneside,margin={-0.6cm,0.8cm},skip=-3pt}
\begin{figure*}[t!]
\centering
\begin{subfigure}[b]{0.45\textwidth}
\includegraphics[height=1.1in]{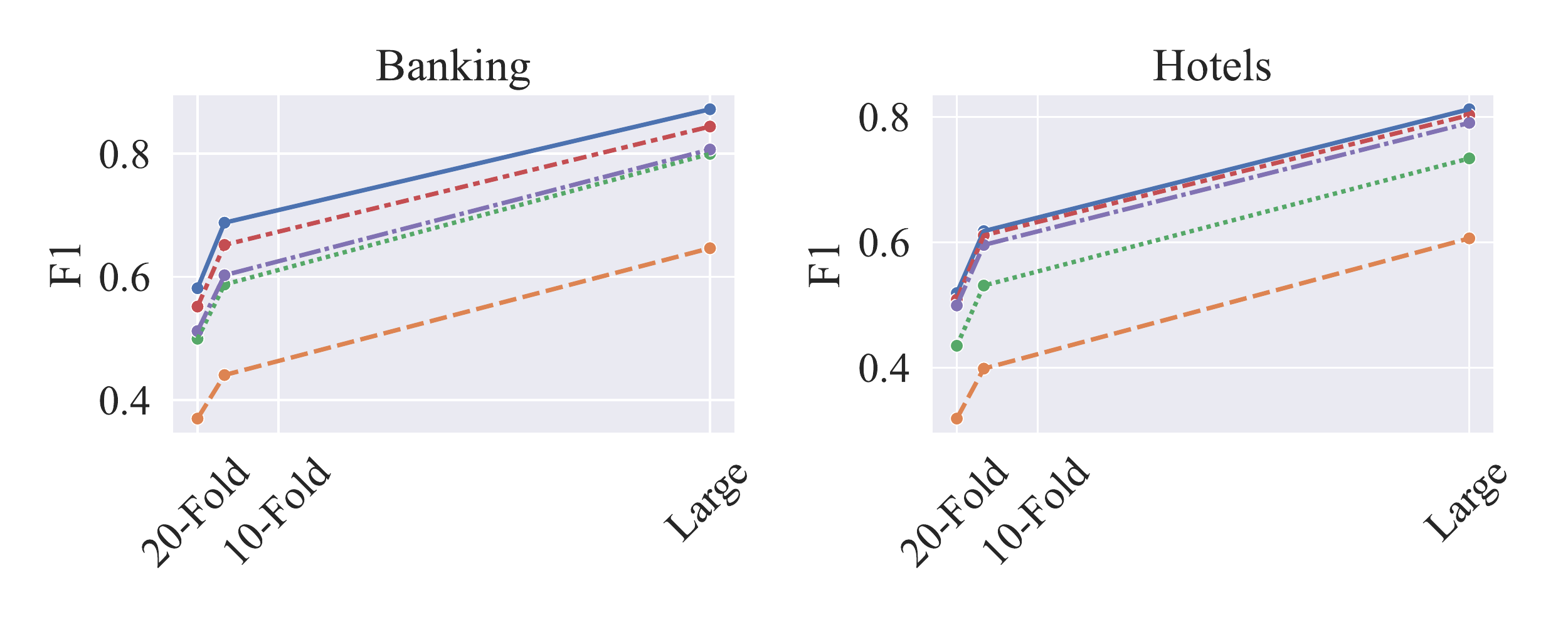}
\caption{\textit{In-domain} in-language results}\label{fig:in-domain cross-lingual}
\end{subfigure}
~
\begin{subfigure}[b]{0.45\textwidth}
\includegraphics[height=1.1in]{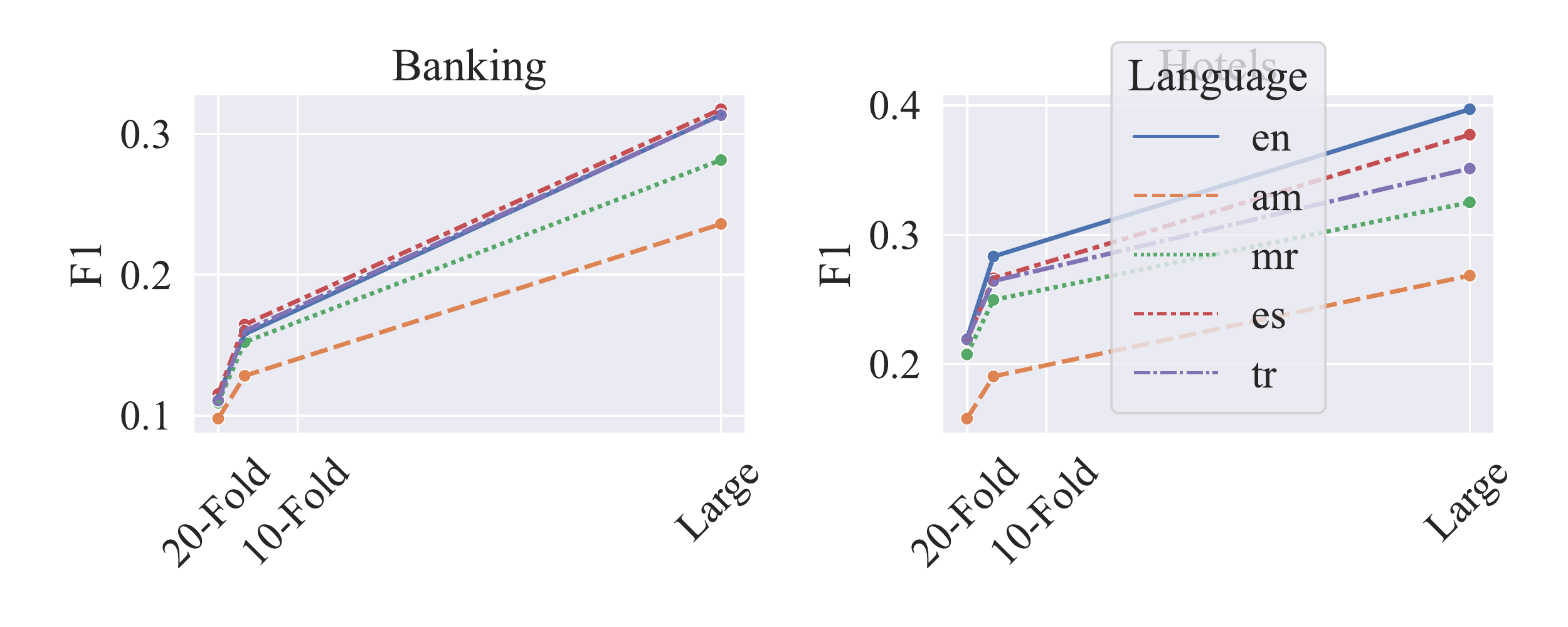}\caption{\textit{Cross-domain in-language} results}\label{fig:cross-domain in-lingual}
\end{subfigure}
~
\begin{subfigure}[b]{0.03\textwidth}
{\raisebox{10mm}{\includegraphics[height=0.7in]{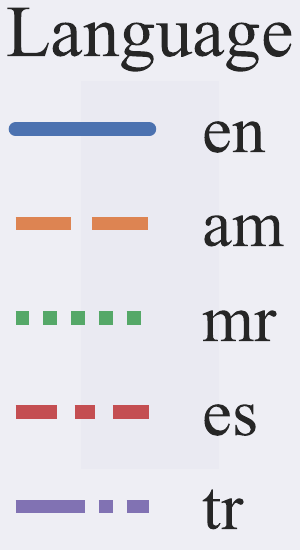}}}
\end{subfigure}
\vspace{-1mm}
\caption{Comparison of in-language \textbf{(a)} in-domain and \textbf{(b)} cross-domain results for intent detection ($F_1$). The model is the \textit{MLP-based} baseline with mpnet as the underlying encoder.}
\vspace{-2mm}
\end{figure*}

\sparagraph{High-Resource and Low-Resource Languages}
    Language models such as mDeBERTa and XLM-R are pretrained on $\sim$100 languages. However, their representational power is uneven for high- and low-resource languages~\citep{lauscher2020zero,ebrahimi-etal-2022-americasnli,wu-etal-2022-zero}. \dataset includes the same training and evaluation data for all languages, allowing us to systematically analyse model performance on dialogue NLU for both high- and low-resource languages. We compare the performance for different languages in the in-domain setup. As seen in Figure~\ref{fig:in-domain cross-lingual}, the overall trend in performance is the same across languages: with more training data, we gain higher performance overall. Interestingly, the absolute numbers are indicative of the resources available in pre-training for a given language. For instance, Amharic (am) has the lowest performance while Spanish (es) has the highest performance.\footnote{This trend holds across domains, sentence encoders, and the zero-shot QA setup. These additional results are provided in Appendix~\ref{app: indomain cross-lingual labse} and Appendix~\ref{app: qa intent results}.}



\rparagraph{Cross-Domain Generalisation} We now consider the intent detection task in the cross-domain in-language setting. 
The results in Figure~\ref{fig:cross-domain in-lingual} corroborate the findings from the in-domain experiments: the lower-resource the language is, the lower the performance on the task is. It is noticeable that the performance in the cross-domain setup is much lower than for the in-domain setup, additionally exposing the complexity of \dataset.  Additionally, in Figure~\ref{fig:cross-domain in-lingual} we observe that for the cross-domain setup, high-resource languages benefit more from the increase in training data size than low-resource languages.  This shows that the gap in performance on low- and high-resource languages is rooted not only in the amount of in-task training data available but also in the representational power of multilingual models for low-resource languages. 

\rparagraph{Future Directions}
Our work focuses on collecting high-quality parallel data through expert translators. With recent advances in MT \citep{kocmi-etal-2022-findings}, it would be worth investigating if the quality of datasets collected with machine translation + human post-editing \citep{hung-etal-2022-multi2woz, ding-etal-2022-globalwoz} is on par with the human translators, especially for the higher resource language pairs.

Another possible direction for future work would be to further diversify the dataset by including additional languages, i.e. with a focus on increasing coverage of language families, branches, and/or scripts, or properties that pose particular challenges in multilingual settings (e.g. free word order in Machine Translation, etc.). 

We believe our resource will have an impact beyond multilingual multi-intent multi-domain systems. We hope the community addresses interesting questions in data augmentation (generating paraphrases with multiple intents), analysing representation learning in multilingual models, translation studies, and MT evaluation.

\section{Conclusion}
We collected \dataset, a dataset that facilitates multilinugual, multi-label, multi-domain NLU for task-oriented dialogue. Our dataset incorporates core properties from its predecessor NLU++ \citep{casanueva-etal-2022-nluplusplus}: a multi-intent and slot-rich ontology, a mixture of generic and domain-specific intents for reusability, and utterances that are based on complex scenarios. We investigated these properties in a multilingual setting for Spanish, Marathi, Turkish, and Amharic. We implemented MLP-based and QA-based baselines for intent detection and slot labelling across different data setups, transfer learning setups, and multilingual models. From a wide set of observations, we highlight that (i) there is a significant drop in performance across all languages as compared to NLU++ with performance drops increasing as we progress from high- to low-resource languages; (ii) zero-shot performance improves when the source language has lower resources in the MLP setup; (iii) cross-lingual transfer in the QA-based intent detection is dependent on matching the script of the source language and amount of data during pretraining setup. 

We hope that the community finds this dataset valuable while working on advancing research in multilingual NLU for task-oriented dialogue. 

\section{Limitations}
\dataset, like NLU++ on which it was based, comprises utterances extracted from real dialogues between users and conversational agents as well as synthetic human-authored utterances constructed with the aim of introducing additional combinations of intents and slots. The utterances, therefore, lack the wider context that would be present in a complete dialogue. As such the dataset cannot be used to evaluate systems with respect to discourse-level phenomena present in dialogue. 

Our source dataset includes scenarios and names that are Anglo-centric and will not capture the nuances of intent detection at the regional and cultural level. Future efforts in collecting multilingual datasets should focus on appropriate localisation of the mentioned domains, intents, and scenarios \citep{liu-etal-2021-visually,majewska2022cross}.

\section{Ethics Statement}
Our dataset builds on the NLU++ dataset \citep{casanueva-etal-2022-nluplusplus} which was collected by removing any personal information. All the names in the dataset are also collected by randomly combining names and surnames from the list of the top 10K names from the US registry.

Our data collection process was thoroughly reviewed by the School of Informatics, University of Edinburgh under the number 2019/59295. Our translators are on legal contracts with the respective translation agencies (See Appendix~\ref{sec:annotation_costs}) for details.

Although we have carefully vetted our datasets to exclude problematic examples, the larger problem of unethical uses and unfairness in conversational systems cannot be neglected \citep{dinan-etal-2021-safety}. Our work also uses multilingual language models that are shown to harm marginalised populations \citep{kaneko-etal-2022-gender}.
Our dataset is publicly available under Creative Commons Attribution 4.0 International (CC-BY-4.0).

\section*{Acknowledgements}
We thank our translators on Proz.com and Blend Express for helping us with the translations in this dataset. We thank Blend Express for offering us 10\% discount on their services. We thank Nina Markl, Stephanie Droop, Laurie Burchell, Ronald Cardenas, Resul Tugay, and Hallelujah Kebede for helping us with our internal evaluation. We are also grateful to I\~{n}igo Casanueva for his feedback and advice while working on the data collection. We thank the anonymous reviewers for their helpful suggestions.

This work was conducted within the
EU project GoURMET H2020–825299. This work was supported in part by the UKRI Centre for Doctoral Training in Natural Language Processing, funded by the UKRI (grant EP/S022481/1) and the University of Edinburgh (Moghe) and by the ERC H2020 Advanced Fellowship GA 742137 SEMANTAX (Guillou). We also thank Huawei for their support (Moghe). The work was also supported in part by a personal Royal Society University Research Fellowship (no 221137; 2022-) awarded to Ivan Vuli\'{c}, as well as by a Huawei research donation to the Language Technology Lab at the University of Cambridge.

\bibliography{anthology,custom}
\bibliographystyle{acl_natbib}

\appendix

\label{sec:appendix}

\section{Language Codes}
\label{app: language codes}
Language codes which are used in the paper are provided in Table \ref{tab: language code}.

\begin{table}[!h]
\small
\centering
\begin{tabular}{@{}lc@{}}
\toprule
Language code & Language \\ \midrule
\textsc{en}            & English  \\
\textsc{am}            & Amharic  \\
\textsc{mr}            & Marathi  \\
\textsc{es}            & Spanish  \\
\textsc{tr}            & Turkish  \\ \bottomrule
\end{tabular}%
\caption{Language codes used in the paper}
\label{tab: language code}
\end{table}

\section{Translation Guidelines}
\label{app:translation_guidelines}
We are an academic team interested in evaluating modern automatic machine translation systems for building better multilingual chatbots. The sentences that you will create should be colloquial. We do not want exact translations but rather what would have been your utterance if you were to say the given content in Spanish. Please do not use any form of machine translation during the process. 

The sentences in this document are spoken to a customer service bot that provides banking services (e.g., making transfers, depositing cheques, reporting lost cards, requesting mortgage information) and hotel ‘bell desk’ reception tasks (e.g., booking rooms, asking about pools or gyms, requesting room service).

Please translate the sentences under ‘source\_text’. Please write the translation under ‘target\_text’. In the translation please:
\begin{itemize}
    \item maintain the meaning and style as close to English text as possible;\\
    Example: ``Exactly, it was declined"\\
    It is important that the colloquial word ``Exactly" is reflected in the translation.
    \item  if the example includes pronouns (e.g., ``this one"), maintain the pronouns in the translation as well.
    \item you are encouraged to translate the proper names and time values in the most natural form for the target language.\\
    Example: ``cancel the one at 2:35 p.m. on Dec 8th" \\
    The ``2:35 p.m." can be translated in any way time is usually expressed in the target language, e.g., ``25 minutes to 3" or ``2:35 in the day", if ``p.m." is non-existent in the target language.
    \item If there is no exact translation for a concept or the concept  is absent from the culture, feel free to substitute it with a description of the concept or a similar concept familiar to the population. 
    \\ Example: ``book a hotel via Booking.com"
    \\ If there is no access to Booking com, feel free to substitute it with ``book a hotel by phone".

\end{itemize}

You may observe that some of the sentences have some spans under slot\_1, slot\_2, slot\_3 and so on. After you have written the Spanish sentence in the target\_text, please replace the values under this column with the corresponding spans in your Spanish sentence in the respective slot columns
\begin{enumerate}
    \item For example, in the sentence ``how much more have I spent on take out since last week?", there is ``last week" under slot\_1 and ``take out" under slot\_2. We would expect you to copy the corresponding phrases (i,e: the translations for last week and take out respectively) from your written Spanish sentence in the ‘slot\_1’ and ‘slot\_2’. If there is no exact phrase that matches the span from English, copy its equivalent in the columns.
    \item Please do not change the order of values while writing the corresponding value columns in the target sentence.
\end{enumerate}

\section{Dataset Collection Details}
\label{sec:annotation_costs}
Our annotators for Spanish are based in Spain, Amharic are based in Ethiopia, Marathi are based in India, and Turkish are based in Turkey.
The rates for translation were fixed by the translators or the translation agencies, ensuring fair pay for the translators. Our internal annotators were compensated with £15/hour for their work. 
We provide details on the dataset collection costs in Table \ref{tab:annotation_breakdown}. 
\begin{table}[!t]
\centering
\small
\begin{tabular}{@{}lllr@{}}
\toprule
\textbf{Type} & \textbf{Platform} & \textbf{Language} & \textbf{Cost (£)} \\ \midrule
Translation & Proz.com & Spanish & 1,500 \\
Translation & BLEND & Marathi & 2,031 \\
Translation & BLEND & Amharic & 2,036 \\
Translation & BLEND & Turkish & 1,954 \\
Internal Evaluation & N/A & All & 103 \\
\midrule
Total &  &  & 7,624 \\ \bottomrule
\end{tabular}
\caption{Dataset collection cost breakdown}
\label{tab:annotation_breakdown}
\end{table}

\section{Full Results for MLP-Based Intent Detection Baselines}
We provide full results for the \textit{MLP-based} baseline in \cref{tab: in-domain in-language,tab: in-domain cross-language labse,tab: in-domain cross-language mpnet,tab: cross-domain in-language,tab: cross-domain cross-language labse,tab: cross-domain cross-language mpnet}. Our implementation uses the Transformers library \citep{huggingface-transformers} and the SentenceBERT library \cite{reimers-2019-sentence-bert}. The models were trained on NVIDIA Titan xP GPUs. Approximate training times for every fold are provided in Table \ref{tab: training time}.

\begin{table}[!h]
\centering
\def\arraystretch{0.8}
\small
\begin{tabular}{@{}l|rrr@{}}
\toprule
\textbf{Domain} & \multicolumn{1}{l}{\textsc{20}} & \multicolumn{1}{l}{\textsc{10}} & \multicolumn{1}{l}{\textsc{Large}} \\ \midrule
\textsc{banking} & 1.5 & 5 & 25 \\
\textsc{hotels} & 1 & 2 & 17 \\
\textsc{all} &  3 & 10.5 & 40  \\ \bottomrule
\end{tabular}
\vspace{-0.5mm}
\caption{Approximate GPU training times for \textit{MLP-based} setup for every fold (in mins).}\label{tab: training time}
\vspace{-1.5mm}
\end{table}
\begin{table}[!h]
\centering
\def\arraystretch{0.8}
\small
\begin{tabular}{@{}l|rrr@{}}
\toprule
\textbf{Domain} & \multicolumn{1}{l}{\textsc{20}} & \multicolumn{1}{l}{\textsc{10}} & \multicolumn{1}{l}{\textsc{Large}} \\ \midrule
\textsc{banking} & 1 & 2 & 8 \\
\textsc{hotels} & 0.33 & 0.75 & 33 \\
\textsc{all} &  2 & 4 & 15  \\ \bottomrule
\end{tabular}
\vspace{-0.5mm}
\caption{Approximate GPU training times for \textit{QA-based} setup for every fold (in hours).}\label{tab: training time qa}
\vspace{-1.5mm}
\end{table}
\label{app: mlp results}

\begin{table*}[!htp]\centering
\scriptsize
\begin{tabularx}{\linewidth}{l YYY YYY YYY}
\toprule
  {} & \multicolumn{3}{c}{\bf \textsc{banking}} & \multicolumn{3}{c}{\bf \textsc{hotels}} & \multicolumn{3}{c}{\bf \textsc{all}} \\
  \cmidrule(lr){2-4} \cmidrule(lr){5-7} \cmidrule(lr){8-10}
\textbf{Model} & \textbf{20-Fold}  & \textbf{10-Fold} & \textbf{Large} & \textbf{20-Fold}  & \textbf{10-Fold} & \textbf{Large} & \textbf{20-Fold}  & \textbf{10-Fold} & \textbf{Large} \\ \midrule
\rowcolor{Gray} {} & \multicolumn{9}{c}{\bf \textsc{English}} \\ \midrule
LaBSE &51.5 &64.3 &87.1 &45.1 &55.4 &80.4 &51.8 &63.4 &85.2\\
mpnet & \textbf{58.1} & \textbf{68.8} & \textbf{87.2} & \textbf{51.9} & \textbf{61.8} & \textbf{81.2} & \textbf{57.7} & \textbf{68.1} & \textbf{86.1} \\\midrule
\rowcolor{Gray} {} & \multicolumn{9}{c}{\bf \textsc{amharic}} \\ \midrule
LaBSE & \textbf{46.2} & \textbf{56.60} & \textbf{77.3} & \textbf{38.6} & \textbf{48.8} & \textbf{71.9} & \textbf{46.4} & \textbf{56.1} & \textbf{76.1} \\
mpnet &37.0 &44.1 &64.7 &31.9 &39.9 &60.6 &36.2 &43.1 &63.5 \\ \midrule
\rowcolor{Gray} {} & \multicolumn{9}{c}{\bf \textsc{marathi}} \\ \midrule
LaBSE & \textbf{50.9} & \textbf{63.5} & \textbf{84.1} &43.5 & \textbf{53.5} & \textbf{77.0} & \textbf{50.8} & \textbf{61.9} & \textbf{81.9} \\
mpnet &50.0 &58.8 &80.0 & \textbf{43.5} &53.1 & 73.4 &49.2 &57.8 &78.0 \\ \midrule
\rowcolor{Gray} {} & \multicolumn{9}{c}{\bf \textsc{spanish}} \\ \midrule
LaBSE &51.4 &64.0 & \textbf{85.2} &46.4 &56.6 &79.5 &52.2 &63.5 & \textbf{83.8} \\
mpnet & \textbf{55.2} & \textbf{65.2} & 84.4 & \textbf{50.9} & \textbf{61.1} & \textbf{80.3} & \textbf{55.4} & \textbf{65.1} &83.5 \\ \midrule
\rowcolor{Gray} {} & \multicolumn{9}{c}{\bf \textsc{turkish}} \\ \midrule
LaBSE &49.2 & \textbf{61.6} & \textbf{82.8} &45.2 &55.9 & \textbf{80.1} &50.1 &61.3 & \textbf{82.0} \\
mpnet & \textbf{51.2} &60.3 &80.7 & \textbf{49.9} & \textbf{59.6} &79.1 & \textbf{52.3} & \textbf{61.3} &80.9 \\
\bottomrule
\end{tabularx}
\caption{In-language in-domain results for intent detection with MLP-based setup ($F_1$ $\times$ 100).}\label{tab: in-domain in-language}
\end{table*}

\begin{table*}[!t]
\def\arraystretch{0.96}
\centering
{\scriptsize
\begin{tabularx}{\linewidth}{ll YYY YYY YYY}
\toprule
  {} & {} & \multicolumn{3}{c}{\bf \textsc{banking}} & \multicolumn{3}{c}{\bf \textsc{hotels}} & \multicolumn{3}{c}{\bf \textsc{all}} \\
  \cmidrule(lr){3-5} \cmidrule(lr){6-8} \cmidrule(lr){9-11}
\textbf{SRC} & \textbf{TGT} & \textbf{20-Fold}  & \textbf{10-Fold} & \textbf{Large} & \textbf{20-Fold}  & \textbf{10-Fold} & \textbf{Large} & \textbf{20-Fold}  & \textbf{10-Fold} & \textbf{Large} \\
\cmidrule(lr){3-11}
\cmidrule(lr){1-11}

\multirow{4}{*}{en} &am &38.9 &49.6 &71.0 &29.8 &38.3 &65.1 &37.7 &47.5 &70.2 \\

&mr &34.1 &47.4 &77.6 &29.2 &38.5 &69.1 &32.8 &45.1 &75.9 \\
&es &36.2 &50.1 &80.8 &34.3 &44.2 &74.8 &35.9 &49.2 &80.3 \\
&tr &29.4 &42.0 &72.9 &27.4 &36.2 &72.0 &28.7 &41.1 &74.2 \\

 \cmidrule(lr){1-11}

\multirow{4}{*}{am} 
&en &48.3 &57.7 &78.8 &43.5 &51.2 &73.7 &48.4 &56.9 &77.8 \\
&mr &38.9 &51.7 &76.0 &33.4 &44.0 &70.1 &39.1 &50.8 &74.4 \\
&es &40.1 &52.6 &76.5 &37.4 &48.1 &73.3 &41.5 &53.0 &76.2 \\
&tr &33.8 &46.5 &73.2 &31.2 &42.2 &70.8 &34.5 &46.4 &73.0 \\

 \cmidrule(lr){1-11}
\multirow{4}{*}{mr} 
&en &51.3 &59.8 &79.4 &47.2 &54.0 &74.6 &49.6 &58.0 &78.7 \\
&am &48.5 &57.1 &71.3 &41.7 &50.2 &69.8 &47.8 &55.8 &70.3 \\
&es &50.0 &61.4 &79.3 &45.5 &55.1 &76.6 &50.8 &60.6 &78.2 \\
&tr &43.8 &55.4 &74.9 &41.5 &50.8 &75.0 &45.2 &55.4 &75.1 \\
 \cmidrule(lr){1-11}

\multirow{4}{*}{es} 
&en &52.2 &60.6 &80.7 &48.7 &55.4 &74.3 &50.9 &59.0 &80.2 \\
&am &47.7 &56.5 &70.7 &40.8 &49.2 &68.8 &47.3 &55.7 &70.1 \\
&mr &47.1 &59.0 &78.9 &40.7 &50.7 &73.3 &47.3 &58.0 &77.2 \\
&tr &43.1 &55.6 &76.5 &39.9 &50.3 &76.2 &44.1 &55.8 &76.7 \\  
\cmidrule(lr){1-11}

\multirow{4}{*}{tr} 
&en &49.7 &56.8 &75.7 &47.3 &53.7 &73.9 &48.0 &55.5 &76.5 \\
&am &48.2 &56.1 &70.6 &42.4 &50.7 &66.8 &47.4 &54.9 &69.5 \\
&mr &49.1 &59.4 &76.6 &44.4 &53.9 &74.1 &49.3 &58.5 &75.4 \\
&es &51.3 &61.5 &78.5 &48.0 &57.6 &78.1 &52.3 &61.5 &78.7 \\
\bottomrule
\end{tabularx}
}%
\caption{Cross-lingual in-domain results for intent detection with MLP-based and LaBSE as the fixed sentence encoder ($F_1$ $\times$ 100).}\label{tab: in-domain cross-language labse}
\end{table*}

\begin{table*}[!t]
\def\arraystretch{0.96}
\centering
{\scriptsize
\begin{tabularx}{\linewidth}{ll YYY YYY YYY}
\toprule
  {} & {} & \multicolumn{3}{c}{\bf \textsc{banking}} & \multicolumn{3}{c}{\bf \textsc{hotels}} & \multicolumn{3}{c}{\bf \textsc{all}} \\
  \cmidrule(lr){3-5} \cmidrule(lr){6-8} \cmidrule(lr){9-11}
\textbf{SRC} & \textbf{TGT} & \textbf{20-Fold}  & \textbf{10-Fold} & \textbf{Large} & \textbf{20-Fold}  & \textbf{10-Fold} & \textbf{Large} & \textbf{20-Fold}  & \textbf{10-Fold} & \textbf{Large} \\
\cmidrule(lr){3-11}
\cmidrule(lr){1-11}

\multirow{4}{*}{en} &am &24.8 &28.6 &32.9 &12.9 &17.5 &23.0 &22.0 &25.6 &30.2 \\
&mr &42.0 &49.6 &63.7 &37.5 &46.1 &62.2 &41.0 &48.3 &61.4 \\
&es &50.1 &59.1 &75.3 &45.8 &55.5 &75.3 &49.2 &58.5 &75.1 \\
&tr &43.9 &52.1 &66.7 &44.8 &54.1 &73.5 &44.5 &52.8 &68.0 \\
 \cmidrule(lr){1-11}

 \multirow{4}{*}{am} &en &37.2 &41.3 &51.2 &35.6 &41.5 &57.4 &36.3 &39.8 &50.8 \\
&mr &37.4 &42.4 &54.2 &34.7 &40.6 &57.1 &36.4 &40.8 &53.5 \\
&es &38.9 &44.5 &56.9 &36.0 &42.7 &60.4 &38.3 &43.1 &55.9 \\
&tr &38.1 &43.5 &55.4 &35.8 &42.3 &60.6 &37.7 &42.4 &55.3 \\ \cmidrule(lr){1-11}

\multirow{4}{*}{mr} &en &51.2 &58.8 &75.5 &48.5 &57.4 &75.4 &51.2 &58.9 &75.9 \\
&am &32.3 &37.4 &47.8 &20.5 &28.1 &43.1 &30.6 &35.2 &46.2 \\
&es &50.3 &58.0 &75.5 &46.7 &56.2 &76.5 &50.3 &58.3 &75.8 \\
&tr &47.2 &54.8 &72.0 &46.5 &55.2 &74.9 &48.2 &55.7 &73.4 \\ \cmidrule(lr){1-11}

\multirow{4}{*}{es} &en &55.9 &64.7 &82.6 &52.4 &61.5 &78.7 &56.2 &64.6 &81.6 \\
&am &30.8 &36.4 &45.5 &16.8 &23.6 &37.5 &28.7 &34.0 &43.1 \\
&mr &46.8 &54.9 &70.4 &41.8 &51.3 &68.2 &46.5 &54.4 &69.4 \\
&tr &49.4 &58.1 &74.8 &48.4 &58.2 &77.9 &50.4 &59.1 &75.6 \\ \cmidrule(lr){1-11}

\multirow{4}{*}{tr} &en &52.8 &60.6 &77.8 &52.3 &61.3 &77.6 &53.6 &61.3 &78.1 \\
&am &32.5 &37.9 &48.1 &18.1 &24.8 &39.5 &30.4 &35.7 &44.6 \\
&mr &46.8 &54.6 &70.9 &42.5 &51.7 &68.1 & 46.5 &54.4 &70.1 \\
&es &52.3 &60.8 &78.7 &49.8 &59.5 &78.3 &53.0 &61.4 &79.3 \\

\bottomrule
\end{tabularx}
}%
\caption{Cross-lingual in-domain results for intent detection with MLP-based and mpnet as the fixed sentence encoder ($F_1$ $\times$ 100).}\label{tab: in-domain cross-language mpnet}

\end{table*}

\begin{table*}[!htp]\centering
\scriptsize
\begin{tabularx}{\linewidth}{l YYY YYY YYY}
\toprule
  {} & \multicolumn{3}{c}{\bf \textsc{banking}} & \multicolumn{3}{c}{\bf \textsc{hotels}} \\
  \cmidrule(lr){2-4} \cmidrule(lr){5-7} 
\textbf{Model} & \textbf{20-Fold}  & \textbf{10-Fold} & \textbf{Large} & \textbf{20-Fold}  & \textbf{10-Fold} & \textbf{Large} \\ \midrule
\rowcolor{Gray} {} & \multicolumn{6}{c}{\bf \textsc{English}} \\ \midrule
LaBSE & 10.6 &15.8 & \textbf{35.2} & \textbf{23.0} & \textbf{29.9} & \textbf{43.4}\\
mpnet & \textbf{11.4} & \textbf{15.8} &31.4 &22.0 &28.3 &39.7 \\\midrule
\rowcolor{Gray} {} & \multicolumn{6}{c}{\bf \textsc{amharic}} \\ \midrule
LaBSE & \phantom{1}8.6 & \textbf{13.6} & \textbf{32.1} & \textbf{22.4} & \textbf{28.5} & \textbf{38.6}\\
mpnet & \phantom{1}\textbf{9.8} &12.8 &23.6 &15.8 &19.0 &26.8 \\ \midrule
\rowcolor{Gray} {} & \multicolumn{6}{c}{\bf \textsc{marathi}} \\ \midrule
LaBSE & \phantom{1}7.2 &12.1 & \textbf{30.1} & \textbf{21.8} & \textbf{28.6} & \textbf{40.5}\\
mpnet & \textbf{10.9} & \textbf{15.2} &28.1 &20.7 &25.0 &32.5 \\ \midrule
\rowcolor{Gray} {} & \multicolumn{6}{c}{\bf \textsc{spanish}} \\ \midrule
LaBSE & \phantom{1}8.2 &13.7 & \textbf{33.9} & \textbf{22.4} & \textbf{29.7} & \textbf{42.0} \\
mpnet & \textbf{11.5} & \textbf{16.5} &31.7 &21.8 &26.6 &37.7 \\ \midrule
\rowcolor{Gray} {} & \multicolumn{6}{c}{\bf \textsc{turkish}} \\ \midrule
LaBSE & \phantom{1}7.6 &12.6 & \textbf{32.8} & \textbf{22.1} & \textbf{28.8} & \textbf{42.2} \\
mpnet & \textbf{11.1} & \textbf{16.0} &31.3 &21.9 &26.4 &35.1 \\
\bottomrule
\end{tabularx}
\caption{In-language cross-domain results for intent detection with MLP-based setup ($F_1$ $\times$ 100).}\label{tab: cross-domain in-language}
\end{table*}

\begin{table*}[!t]
\def\arraystretch{0.96}
\centering
{\scriptsize
\begin{tabularx}{\linewidth}{ll YYY YYY}
\toprule
  {} & {} & \multicolumn{3}{c}{\bf \textsc{banking}} & \multicolumn{3}{c}{\bf \textsc{hotels}}\\
  \cmidrule(lr){3-5} \cmidrule(lr){6-8} 
\textbf{SRC} & \textbf{TGT} & \textbf{20-Fold}  & \textbf{10-Fold} & \textbf{Large} & \textbf{20-Fold}  & \textbf{10-Fold} & \textbf{Large}\\
\cmidrule(lr){3-8}
\cmidrule(lr){1-8}
\multirow{4}{*}{en}
&am & \phantom{1}6.8 &10.4 &29.0 &19.3 &25.9 &40.8 \\

&mr & \phantom{1}3.8 & \phantom{1}6.1 &23.9 &15.1 &20.8 &39.1 \\
&es & \phantom{1}4.3 & \phantom{1}7.2 &26.4 &16.0 &22.5 &41.0 \\
&tr & \phantom{1}2.7 & \phantom{1}4.5 &22.7 &13.3 &19.2 &40.8 \\
\cmidrule(lr){1-8}

\multirow{4}{*}{am} 
&en &12.4 &18.0 &33.5 &22.6 &27.2 &37.5 \\
&mr & \phantom{1}4.7 & \phantom{1}8.92 &27.14 &18.54 &25.04 &36.05 \\
&es & \phantom{1}5.5 &10.4 &29.0 &19.3 &26.0 &38.1 \\
&tr & \phantom{1}3.4 & \phantom{1}6.9 &25.2 &16.1 &23.3 &38.8 \\  \cmidrule(lr){1-8}

\multirow{4}{*}{mr} 
&en &14.1 &19.3 &33.1 &24.4 &29.0 &38.2 \\
&am &11.0 &16.3 &31.5 &24.1 &29.7 &39.3 \\
&es & \phantom{1}8.1 &13.5 &30.5 &22.9 &29.3 &39.9 \\
&tr & \phantom{1}5.5 & \phantom{1}9.9 &27.9 &19.4 &26.6 &40.8 \\ \cmidrule(lr){1-8}

\multirow{4}{*}{es} 
&en &14.6 &19.6 &32.8 &25.0 &29.3 &38.6 \\
&am &11.0 &16.4 &32.8 &24.0 &30.0 &40.3 \\
&mr & \phantom{1}7.0 &11.7 &32.1 &20.6 &27.4 &40.7 \\
&tr & \phantom{1}5.3 & \phantom{1}9.6 &29.6 &18.9 &26.7 &41.3 \\ \cmidrule(lr){1-8}

\multirow{4}{*}{tr} 
&en &15.9 &20.5 &34.0 &24.0 &27.8 &36.8 \\
&am &13.3 &18.6 &33.8 &24.4 &29.2 &38.7 \\
&mr & \phantom{1}9.3 &15.0 &34.1 &22.4 &28.3 &38.7 \\
&es &10.5 &16.7 &35.4 &23.9 &30.0 &40.4 \\

\bottomrule
\end{tabularx}
}%
\caption{Cross-lingual cross-domain results for intent detection with MLP-based and LaBSE as the fixed sentence encoder ($F_1$ $\times$ 100).}\label{tab: cross-domain cross-language labse}
\end{table*}

\begin{table*}[!t]
\def\arraystretch{0.96}
\centering
{\scriptsize
\begin{tabularx}{\linewidth}{ll YYY YYY}
\toprule
  {} & {} & \multicolumn{3}{c}{\bf \textsc{banking}} & \multicolumn{3}{c}{\bf \textsc{hotels}}\\
  \cmidrule(lr){3-5} \cmidrule(lr){6-8} 
\textbf{SRC} & \textbf{TGT} & \textbf{20-Fold}  & \textbf{10-Fold} & \textbf{Large} & \textbf{20-Fold}  & \textbf{10-Fold} & \textbf{Large}\\
\cmidrule(lr){3-8}
\cmidrule(lr){1-8}

\multirow{4}{*}{en} 
&am &  \phantom{1}6.5 &  \phantom{1}8.7 &16.2 &11.2 &14.4 &18.5 \\

&mr &  \phantom{1}9.5 &13.1 &25.9 &18.9 &24.1 &31.9 \\
&es &  \phantom{1}9.9 &13.6 &28.4 &20.1 &25.4 &35.9 \\
&tr &  \phantom{1}9.2 &12.9 &25.9 &19.7 &25.3 &33.5 \\
\cmidrule(lr){1-8}

\multirow{4}{*}{am} 
&en &11.9 &14.3 &23.1 &15.7 &17.2 &18.9 \\
&mr &11.4 &14.4 &24.2 &17.0 &18.9 &22.2 \\
&es &11.7 &14.3 &24.8 &16.7 &18.5 &21.5 \\
&tr &11.3 &14.0 &25.0 &16.9 &19.0 &22.3 \\
\cmidrule(lr){1-8}

\multirow{4}{*}{mr} 
&en &11.9 &16.1 &27.4 &20.7 &24.0 &30.3 \\

&am &  \phantom{1}7.9 &10.9 &21.5 &15.1 &18.9 &25.7 \\
&es &11.0 &15.4 &28.6 &20.8 &24.8 &32.3 \\
&tr &10.6 &15.0 &27.6 &20.9 &24.8 &32.3 \\
\cmidrule(lr){1-8}

\multirow{4}{*}{es} 
&en &12.3 &17.1 &30.9 &22.0 &26.0 &36.0 \\
&am &  \phantom{1}7.4 &11.0 &23.2 &15.0 &19.7 &28.3 \\
&mr &10.8 &15.6 &30.4 &21.2 &26.1 &35.8 \\
&tr &10.7 &15.7 &30.3 &21.8 &26.7 &36.8 \\ 
\cmidrule(lr){1-8}

\multirow{4}{*}{tr} 
&en &12.6 &17.3 &30.7 &21.5 &25.4 &32.2 \\
&am &  \phantom{1}8.1 &11.6 &24.3 &15.6 &19.7 &27.3 \\
&mr &11.1 &15.8 &30.9 &21.2 &25.4 &33.8 \\
&es &11.6 &16.3 &31.7 &21.8 &26.1 &33.7 \\

\bottomrule
\end{tabularx}
}%
\caption{Cross-lingual cross-domain results for intent detection with MLP-based and mpnet as the fixed sentence encoder ($F_1$ $\times$ 100).}\label{tab: cross-domain cross-language mpnet}
\end{table*}

\section{In-Domain Cross-Lingual Results}
\label{app: indomain cross-lingual labse}
\begin{figure*}[t!]
    \centering   \includegraphics[width=0.95\textwidth]{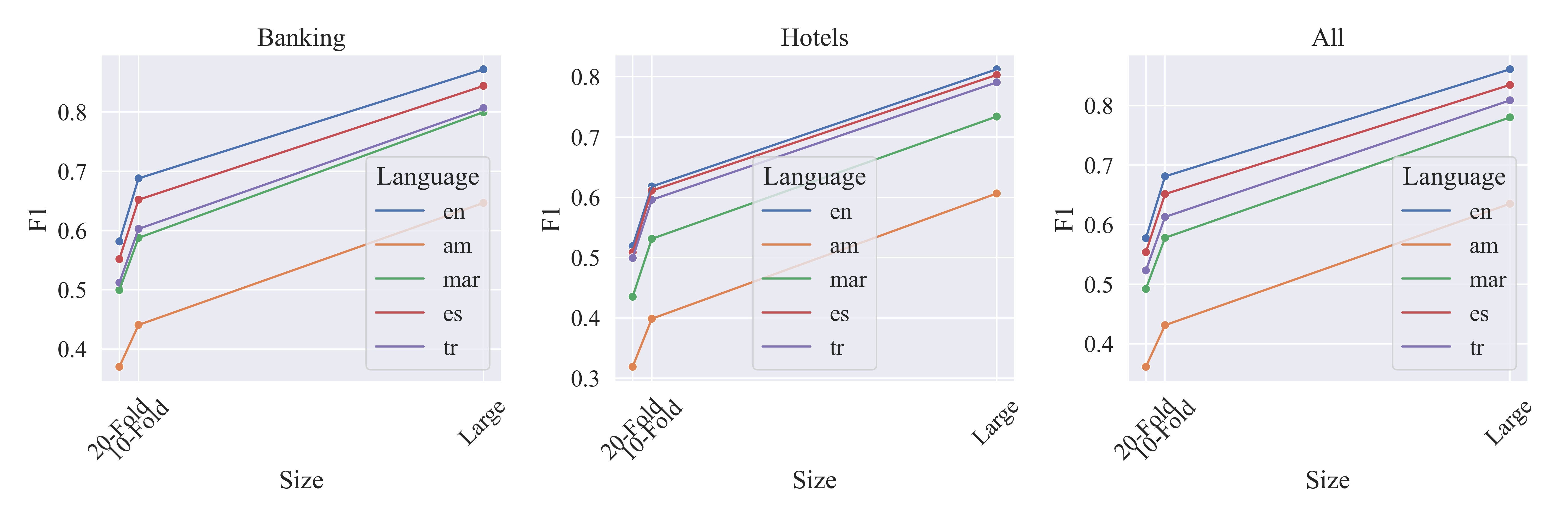}
    \caption{Comparison of in-domain in-lingual results. The results are for \textit{MLP-based} baseline with mpnet as the underlying encoder. English is used as the source language in the experiments.}\label{fig: in-domain cross-lingual mpnet trends}
\end{figure*}

\begin{figure*}[t!]
    \centering   \includegraphics[width=0.95\textwidth]{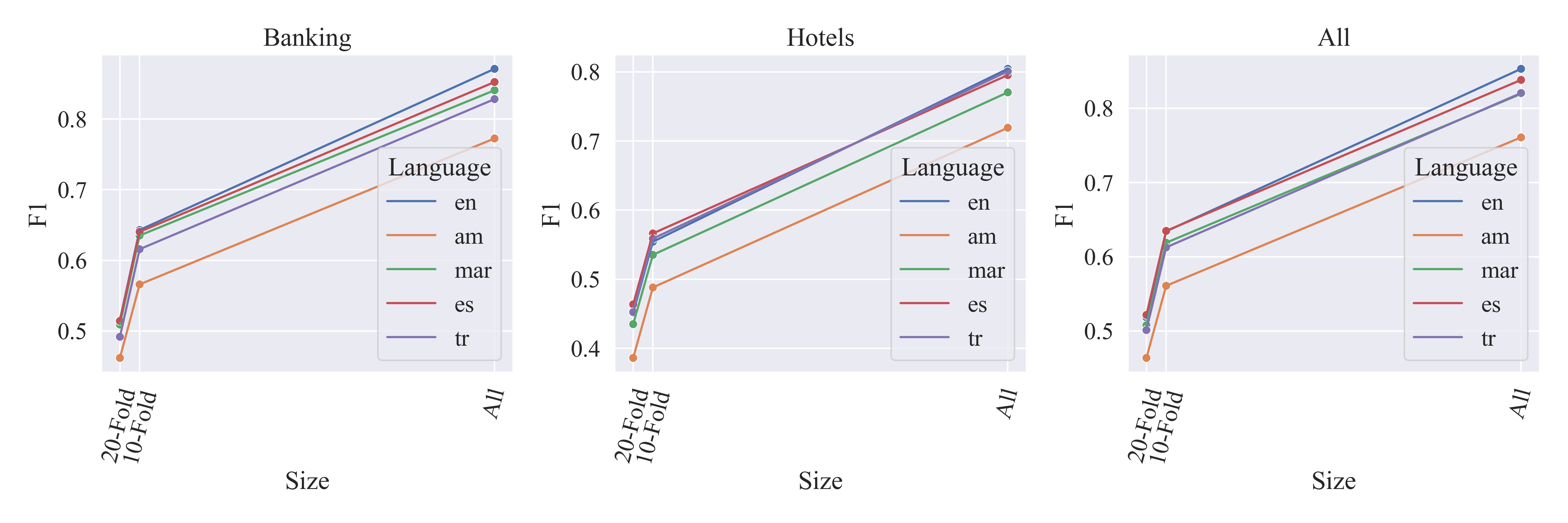}
    \caption{Comparison of in-domain in-lingual results. The results are for \textit{MLP-based} baseline with LaBSE as the underlying encoder. English is used as the source language in the experiments.}\label{fig: in-domain cross-lingual labse trends}
\end{figure*}

We compare the in-domain cross-lingual results using mpnet and LaBSE as an underlying encoder for all domains in Figures~\ref{fig: in-domain cross-lingual mpnet trends} and \ref{fig: in-domain cross-lingual labse trends}, respectively.

\section{Full Results for QA-Based Intent Detection}
\label{app: qa intent results}
We provide full results for the QA-based baseline in \cref{tab: in-domain in-language qa,tab: in-domain cross-language qa,tab: in-domain translate-test qa,tab: cross-domain in-language qa,tab: cross-domain cross-language qa}.

For both slot labelling and intent detection, we performed the hyperparameter search over fold 0 of the \textsc{hotels} domain in the 20-fold setup. The hyperparameters varied include learning rate [1e-5, 2e-5, 3e-5], batch size [4,8,16], epochs [5,10,15] and multilingual models of XLM-R and mDeBERTa. The other hyperparameters are the same as \citep{casanueva-etal-2022-nluplusplus}. Approximate training times for every fold on NVIDIA GeForce RTX 2080 are provided in Table \ref{tab: training time qa}.
Our implementation of the QA model uses the QA fine-tuning scripts from the Transformers Library \citep{huggingface-transformers}.

\begin{table*}[!htp]\centering
\scriptsize
\begin{tabularx}{\linewidth}{l YYY YYY YYY}
\toprule
  {} & \multicolumn{3}{c}{\bf \textsc{banking}} & \multicolumn{3}{c}{\bf \textsc{hotels}} & \multicolumn{3}{c}{\bf \textsc{all}} \\
  \cmidrule(lr){2-4} \cmidrule(lr){5-7} \cmidrule(lr){8-10}
\textbf{Model} & \textbf{20-Fold}  & \textbf{10-Fold} & \textbf{Large} & \textbf{20-Fold}  & \textbf{10-Fold} & \textbf{Large} & \textbf{20-Fold}  & \textbf{10-Fold} & \textbf{Large} \\ \midrule
\rowcolor{Gray} {} & \multicolumn{9}{c}{\bf \textsc{English}} \\ \midrule
mDeBERTa & 80.8 & 85.0 & 93.2 & 66.9 & 73.2 & 87.3 & 79.6 & 83.6 & 91.7 \\ \midrule
\rowcolor{Gray} {} & \multicolumn{9}{c}{\bf \textsc{amharic}} \\ \midrule
mDeBERTa & 56.1 & 64.8 & 83.4 & 51.1 & 57.3 & 77.9 & 57.2 & 64.2 & 81.8 \\ \midrule
\rowcolor{Gray} {} & \multicolumn{9}{c}{\bf \textsc{marathi}} \\ \midrule
mDeBERTa & 70.4 & 77.4 & 88.6 & 53.2 & 63.2 & 81.7 & 69.1 & 75.5 & 86.8 \\ \midrule
\rowcolor{Gray} {} & \multicolumn{9}{c}{\bf \textsc{spanish}} \\ \midrule
mDeBERTa & 78.8 & 83.5 & 91.6 & 65.9 & 72.7 & 86.0 & 77.7 & 82.1 & 89.9 \\ \midrule
\rowcolor{Gray} {} & \multicolumn{9}{c}{\bf \textsc{turkish}} \\ \midrule
mDeBERTa & 75.8 & 80.2 & 90.1 & 65.6 & 72.8 & 86.7 & 75.0 & 79.5 & 89.2 \\ 
\bottomrule
\end{tabularx}
\caption{In-language in-domain results for intent detection with QA-based setup and mDeBERTa as the base model ($F_1$ $\times$ 100).}\label{tab: in-domain in-language qa}
\end{table*}

\begin{table*}[!t]
\def\arraystretch{0.96}
\centering
{\scriptsize
\begin{tabularx}{\linewidth}{ll YYY YYY YYY}
\toprule
  {} & {} & \multicolumn{3}{c}{\bf \textsc{banking}} & \multicolumn{3}{c}{\bf \textsc{hotels}} & \multicolumn{3}{c}{\bf \textsc{all}} \\
  \cmidrule(lr){3-5} \cmidrule(lr){6-8} \cmidrule(lr){9-11}
\textbf{SRC} & \textbf{TGT} & \textbf{20-Fold}  & \textbf{10-Fold} & \textbf{Large} & \textbf{20-Fold}  & \textbf{10-Fold} & \textbf{Large} & \textbf{20-Fold}  & \textbf{10-Fold} & \textbf{Large} \\
\cmidrule(lr){3-11}
\cmidrule(lr){1-11}

\multirow{4}{*}{en} & am & 42.7 & 46.7 & 52.1 & 37.1 & 38.0 & 45.2 & 40.7 & 41.2 & 49.5 \\
& mr & 58.4 & 63.2 & 71.0 & 51.1 & 55.6 & 66.0 & 56.5 & 59.1 & 66.6 \\
& es & 66.2 & 71.3 & 82.4 & 62.2 & 69.1 & 84.1 & 67.9 & 71.9 & 81.2 \\
& tr & 56.5 & 62.2 & 73.0 & 59.6 & 65.3 & 79.6 & 58.2 & 62.1 & 73.6 \\

 \cmidrule(lr){1-11}

\multirow{4}{*}{am}& en & 30.1 & 37.9 & 55.0 & 30.5 & 38.6 & 60.0 & 29.4 & 32.4 & 53.1 \\
& mr & 31.3 & 40.6 & 53.7 & 27.5 & 34.2 & 55.3 & 32.3 & 35.4 & 47.8 \\
& es & 27.9 & 35.5 & 53.1 & 31.5 & 39.7 & 61.2 & 27.7 & 31.2 & 52.2 \\
& tr & 28.5 & 38.2 & 54.3 & 33.0 & 41.3 & 62.5 & 30.7 & 35.0 & 52.4 \\

\cmidrule(lr){1-11}

\multirow{4}{*}{mr}& en & 49.7 & 55.0 & 69.2 & 44.3 & 52.4 & 68.0 & 47.9 & 51.3 & 66.6 \\
& am & 32.4 & 35.6 & 48.4 & 27.1 & 31.0 & 44.4 & 30.4 & 34.0 & 43.5 \\
& es & 46.4 & 50.7 & 65.8 & 43.8 & 51.5 & 66.2 & 45.6 & 49.4 & 63.7 \\
& tr & 44.9 & 50.1 & 62.9 & 44.3 & 52.2 & 67.6 & 45.4 & 49.5 & 61.5 \\

\cmidrule(lr){1-11}

\multirow{4}{*}{es}& en & 67.6 & 74.1 & 84.6 & 62.9 & 69.1 & 84.6 & 67.9 & 73.9 & 84.2 \\
& am & 41.6 & 45.4 & 53.7 & 35.7 & 38.1 & 47.5 & 40.1 & 43.6 & 51.2 \\
& mr & 54.1 & 60.2 & 66.7 & 49.6 & 53.8 & 62.6 & 52.8 & 56.6 & 62.7 \\
& tr & 52.3 & 58.9 & 70.5 & 58.1 & 63.5 & 78.5 & 55.7 & 61.4 & 71.7 \\

\cmidrule(lr){1-11}

\multirow{4}{*}{tr} &en &52.78 &60.57 &77.78 &52.27 &61.25 &77.63 &53.61 &61.25 &78.13 \\
&am &32.46 &37.88 &48.12 &18.05 &24.75 &39.54 &30.37 &35.72 &44.58 \\
&mr &46.83 &54.59 &70.90 &42.46 &51.65 &68.06 &46.5 &54.35 &70.11 \\
&es &52.30 &60.83 &78.73 &49.76 &59.48 &78.25 &52.98 &61.44 &79.27 \\

\bottomrule
\end{tabularx}
}%
\caption{Cross-lingual in-domain results for intent detection with QA-based setup and mDeBERTa as the base model ($F_1$ $\times$ 100).}\label{tab: in-domain cross-language qa}
\end{table*}

\begin{table*}[!t]
\def\arraystretch{0.96}
\centering
{\scriptsize
\begin{tabularx}{\linewidth}{ll YYY YYY YYY}
\toprule
  {} & {} & \multicolumn{3}{c}{\bf \textsc{banking}} & \multicolumn{3}{c}{\bf \textsc{hotels}} & \multicolumn{3}{c}{\bf \textsc{all}} \\
  \cmidrule(lr){3-5} \cmidrule(lr){6-8} \cmidrule(lr){9-11}
\textbf{SRC} & \textbf{TGT} & \textbf{20-Fold}  & \textbf{10-Fold} & \textbf{Large} & \textbf{20-Fold}  & \textbf{10-Fold} & \textbf{Large} & \textbf{20-Fold}  & \textbf{10-Fold} & \textbf{Large} \\
\cmidrule(lr){3-11}
\cmidrule(lr){1-11}

\multirow{4}{*}{en} & am & 17.6 & 21.4 & 25.4 & 11.5 & 13.7 & 20.2 & 16.2 & 18.9 & 23.9 \\
 & mr & 63.5 & 68.1 & 75.8 & 57.2 & 62.7 & 76.8 & 63.3 & 67.8 & 75.4 \\
 & es & 67.7 & 72.3 & 81.1 & 59.9 & 66.5 & 80.9 & 67.8 & 71.6 & 79.8 \\
 & tr & 45.3 & 52.3 & 63.8 & 42.6 & 47.4 & 67.3 & 46.3 & 52.1 & 64.1 \\
 \cmidrule(lr){1-11}

\multirow{4}{*}{am} & en & 20.2 & 24.0 & 32.9 & 12.2 & 13.2 & 21.2 & 16.2 & 19.3 & 28.1 \\
 & mr & 18.8 & 22.2 & 31.2 & 11.7 & 12.4 & 18.2 & 15.2 & 18.1 & 27.7 \\
 & es & 14.2 & 16.6 & 24.1 & 11.4 & 12.3 & 21.5 & 11.7 & 13.7 & 22.3 \\
 & tr & 13.8 & 15.3 & 20.7 & 11.8 & 12.5 & 21.3 & 11.7 & 13.3 & 19.5 \\

\bottomrule
\end{tabularx}
}%
\caption{Translate-test in-domain results for intent detection with QA-based setup and mDeBERTa as the base model ($F_1$ $\times$ 100).}\label{tab: in-domain translate-test qa}
\end{table*}

\begin{table*}[!htp]\centering
\scriptsize
\begin{tabularx}{\linewidth}{l YYY YYY }
\toprule
  {} & \multicolumn{3}{c}{\bf \textsc{banking}} & \multicolumn{3}{c}{\bf \textsc{hotels}}   \\
  \cmidrule(lr){2-4} \cmidrule(lr){5-7} 

\textbf{Model} & \textbf{20-Fold}  & \textbf{10-Fold} & \textbf{Large} & \textbf{20-Fold}  & \textbf{10-Fold} & \textbf{Large}\\ \midrule

\rowcolor{Gray} {} & \multicolumn{6}{c}{\bf \textsc{English}} \\ \midrule
mDeBERTa & 58.3 & 63.6 & 72.8 & 55.9 & 59.0 & 61.8 \\\midrule
\rowcolor{Gray} {} & \multicolumn{6}{c}{\bf \textsc{amharic}} \\ \midrule
mDeBERTa & 41.8 & 47.0 & 56.7 & 35.6 & 39.5 & 49.0 \\\midrule
\rowcolor{Gray} {} & \multicolumn{6}{c}{\bf \textsc{marathi}} \\ \midrule
mDeBERTa & 47.0 & 51.1 & 59.3 & 42.3 & 46.5 & 58.0 \\\midrule
\rowcolor{Gray} {} & \multicolumn{6}{c}{\bf \textsc{spanish}} \\ \midrule
mDeBERTa & 62.1 & 66.6 & 72.0 & 51.3 & 56.6 & 60.1  \\ \midrule
\rowcolor{Gray} {} & \multicolumn{6}{c}{\bf \textsc{turkish}} \\ \midrule
mDeBERTa & 61.6 & 64.9 & 70.9 & 47.6 & 52.6 & 58.7  \\
\bottomrule
\end{tabularx}
\caption{In-language cross-domain results for intent detection with QA-based setup and mDeBERTa as the base model ($F_1$ $\times$ 100).}\label{tab: cross-domain in-language qa}
\end{table*}

\begin{table*}[!t]
\def\arraystretch{0.96}
\centering
{\scriptsize
\begin{tabularx}{\linewidth}{ll YYY YYY}
\toprule
  {} & {} & \multicolumn{3}{c}{\bf \textsc{banking}} & \multicolumn{3}{c}{\bf \textsc{hotels}}\\
  \cmidrule(lr){3-5} \cmidrule(lr){6-8} 
\textbf{SRC} & \textbf{TGT} & \textbf{20-Fold}  & \textbf{10-Fold} & \textbf{Large} & \textbf{20-Fold}  & \textbf{10-Fold} & \textbf{Large}\\
\cmidrule(lr){3-8}
\cmidrule(lr){1-8}

\multirow{4}{*}{en} & am  & 29.0 & 33.7 & 40.3 & 30.5 & 33.9 & 42.6 \\
                    & mar & 45.0 & 50.0 & 56.6 & 40.4 & 44.4 & 52.7 \\
                    & sp  & 54.2 & 61.5 & 71.6 & 49.0 & 52.0 & 59.2 \\
                    & tr  & 52.1 & 59.0 & 68.1 & 42.8 & 46.1 & 53.8 \\ \cmidrule(lr){1-8}
\multirow{4}{*}{am} & en  & 19.7 & 27.7 & 36.3 & 19.4 & 22.1 & 33.1 \\
                    & mr  & 19.6 & 27.7 & 35.5 & 19.6 & 23.6 & 35.6 \\
                    & sp  & 18.9 & 26.5 & 34.6 & 17.8 & 19.0 & 30.5 \\
                    & tr  & 20.8 & 31.5 & 38.2 & 16.9 & 19.5 & 32.7 \\ \cmidrule(lr){1-8}
\multirow{4}{*}{mr} & en  & 36.3 & 40.7 & 51.7 & 35.5 & 38.5 & 46.9 \\
                    & am  & 20.5 & 24.8 & 35.6 & 21.7 & 25.4 & 34.6 \\
                    & sp  & 37.5 & 40.5 & 51.9 & 34.5 & 37.5 & 46.9 \\
                    & tr  & 36.5 & 41.0 & 50.0 & 31.6 & 35.8 & 43.4 \\ \cmidrule(lr){1-8}
\multirow{4}{*}{es} & en  & 60.4 & 65.4 & 71.1 & 46.3 & 50.9 & 53.6 \\
                    & am  & 33.3 & 35.5 & 41.1 & 27.0 & 29.4 & 36.8 \\
                    & mr  & 47.5 & 50.2 & 53.4 & 34.7 & 39.2 & 45.6 \\
                    & tr  & 55.9 & 60.5 & 66.7 & 36.3 & 40.7 & 46.9 \\ \cmidrule(lr){1-8}
\multirow{4}{*}{tr} & en  & 57.0 & 61.0 & 71.0 & 42.6 & 47.5 & 55.6 \\
                    & am  & 33.2 & 36.0 & 43.8 & 25.7 & 29.8 & 38.9 \\
                    & mr  & 46.9 & 50.2 & 57.2 & 34.1 & 39.9 & 48.9 \\ 
                    & es  & 56.3 & 60.6 & 69.0 & 38.2 & 42.6 & 53.0 \\ \bottomrule

\end{tabularx}
}%
\caption{Cross-lingual cross-domain results for intent detection with QA-based setup and mDeBERTa as the base model ($F_1$ $\times$ 100).}\label{tab: cross-domain cross-language qa}
\end{table*}

\section{Full Results for Slot Labelling Baselines}
\label{app: slot labeling results}
We provide full results for slot labelling for \textit{token classification} setup in Tables \ref{tab: token classification in-domain in-language slot labeling} and \ref{tab: token classification  cross-lingual slot labelling}. The results for the \textit{QA-based} setup are provided in Tables \cref{tab: in-domain in-language slot,tab: qa based  cross-lingual slot labelling}.
\begin{table*}[!htp]\centering
\scriptsize
\begin{tabularx}{\linewidth}{l YYY YYY YYY}
\toprule
  {} & \multicolumn{3}{c}{\bf \textsc{banking}} & \multicolumn{3}{c}{\bf \textsc{hotels}} & \multicolumn{3}{c}{\bf \textsc{all}} \\
  \cmidrule(lr){2-4} \cmidrule(lr){5-7} \cmidrule(lr){8-10}
\textbf{Model} & \textbf{20-Fold}  & \textbf{10-Fold} & \textbf{Large} & \textbf{20-Fold}  & \textbf{10-Fold} & \textbf{Large} & \textbf{20-Fold}  & \textbf{10-Fold} & \textbf{Large} \\ \midrule
\rowcolor{Gray} {} & \multicolumn{9}{c}{\bf \textsc{English}} \\ \midrule
XLM-R &37.2 & 59.7 & 85.2 & 24.2  & 54.5 & 79.0 & 43.7 & 62.9 & 83.2\\\midrule
\rowcolor{Gray} {} & \multicolumn{9}{c}{\bf \textsc{amharic}} \\ \midrule
XLM-R & 27.2 & 44.9 & 69.3 & 20.0 & 34.6 & 65.2  & 33.3 & 44.4 & 68.2 \\\midrule
\rowcolor{Gray} {} & \multicolumn{9}{c}{\bf \textsc{marathi}} \\ \midrule
XLM-R & 35.1 & 46.4 &  71.0 & 24.9 & 37.9 &  62.2 & 31.6 & 46.1 & 67.8 \\\midrule
\rowcolor{Gray} {} & \multicolumn{9}{c}{\bf \textsc{spanish}} \\ \midrule
XLM-R & 36.5 & 47.7 & 69.9 & 29.5 & 44.5 & 65.0  & 36.6 & 51.72 & 68.9 \\ \midrule
\rowcolor{Gray} {} & \multicolumn{9}{c}{\bf \textsc{turkish}} \\ \midrule
XLM-R & 32.2 & 52.0 & 82.1 & 19.3 & 49.0 & 71.2  & 38.2 & 55.7 & 76.6 \\
\bottomrule
\end{tabularx}
\caption{In-domain in-language slot labelling performance using XLM-Roberta ($F_1$ $\times$ 100)}\label{tab: token classification in-domain in-language slot labeling}
\end{table*}

\begin{table*}[!t]
\def\arraystretch{0.96}
\centering
{\scriptsize
\begin{tabularx}{\linewidth}{ll YYY YYY YYY}
\toprule
  {} & {} & \multicolumn{3}{c}{\bf \textsc{banking}} & \multicolumn{3}{c}{\bf \textsc{hotels}} & \multicolumn{3}{c}{\bf \textsc{all}}\\
  \cmidrule(lr){3-5} \cmidrule(lr){6-8} \cmidrule(lr){9-11} 
\textbf{SRC} & \textbf{TGT} & \textbf{20-Fold}  & \textbf{10-Fold} & \textbf{Large} & \textbf{20-Fold}  & \textbf{10-Fold} & \textbf{Large} & \textbf{20-Fold}  & \textbf{10-Fold} & \textbf{Large}\\
\cmidrule(lr){3-11}
\cmidrule(lr){1-11}

\multirow{5}{*}{en} 
&en & \phantom{1}37.2  & 59.7 & 85.2  & 24.2 & 54.5 &   79.0 & 43.6 & 62.9 & 83.2 \\ 
&am & \phantom{1}15.2  & 27.1 & 41.2  &  6.1 & 21.2 &   31.2 & 16.4 & 26.8 & 37.0 \\
&mr & \phantom{1}19.7  & 33.4 & 50.5  & 10.8 & 29.5 &   40.9 & 21.9 & 34.0 & 46.8 \\
&es & \phantom{1}22.8  & 29.6 & 58.5  & 14.2 & 39.4 &   56.1 & 28.3 & 43.2 & 58.8 \\
&tr & \phantom{1}18.7  & 35.4 & 56.9  & 14.3 & 37.8 &   49.8 & 26.3 & 41.0 & 54.7 \\

 \cmidrule(lr){1-11}
 \multirow{5}{*}{am} 
&am & 27.2 & 44.8 & 69.2 & 20.0 & 34.5 & 65.2  & 33.2 & 44.3 & 68.2 \\ 
&en & 19.8 & 30.4 & 44.5 & 7.9  & 25.0 & 39.4  & 19.9 & 28.3 & 43.8 \\
&mr & \phantom{1}22.8 & 38.3 & 60.5 & 6.3 & 26.1 &  45.1 & 21.8 & 34.0 & 53.0 \\
&es & \phantom{1}15.5 & 25.6 & 36.5 & 5.5 & 19.5 &  33.9 & 15.4 & 23.6 & 36.6 \\
&tr & \phantom{1}20.3 & 35.7 & 50.6 & 6.7 & 27.3 &  43.3 & 21.5 & 31.8 & 47.7 \\  
\cmidrule(lr){1-11}

\multirow{5}{*}{mr}  
&mr & 35.1 & 46.4 & 71.0 & 24.9 & 37.9 & 62.2  & 31.6 & 46.1 & 67.7  \\
&en & 19.5 & 34.7 & 46.9 & 12.4 & 29.5 & 46.3  & 24.7 & 35.3 & 48.4 \\
&am & 18.6 & 30.5 & 60.0 &  6.7 & 20.9 & 42.8  & 19.2 & 31.3 & 47.5 \\
&es & \phantom{1}13.8 & 25.9 & 37.0 & 08.1 & 21.9 &  36.5 & 17.0 & 26.2 & 37.1 \\
&tr & \phantom{1}20.3 & 35.2 & 50.9 & 11.6 & 32.5 &  47.0 & 25.7 & 37.1 & 50.5 \\ 
\cmidrule(lr){1-11}

\multirow{5}{*}{es} 
&es & 36.5 & 47.6 & 69.8 & 29.5 & 44.4 & 65.0  & 36.6 & 51.7 & 68.8 \\ 
&en & 33.5 & 51.0 & 73.0 & 21.9 & 46.3 & 60.3  & 37.4 & 53.3 & 69.4 \\
&am & 16.6 & 28.6 & 43.2 &  7.8 & 24.0 & 31.7  & 19.1 & 30.4 & 37.5\\

&mr & \phantom{1}21.1 & 32.3  & 51.0 & 12.7 & 31.7 & 41.4  & 24.3 & 35.3&  45.1\\
&tr & \phantom{1}20.0 & 33.9  & 55.4 & 15.6 & 36.7 & 45.6  & 27.1 & 40.5&  50.9 \\ 
\cmidrule(lr){1-11}

\multirow{5}{*}{tr} 
&tr & 32.2 & 52.0 & 82.0 & 19.3 & 49.0 & 71.2  & 38.2 & 55.6 & 76.6 \\
&en & 27.4 & 43.9 & 63.9 & 15.0 & 36.4 & 55.5  & 29.3 & 43.5 & 58.8 \\
&am & 21.6 & 34.1 & 48.0 & 07.1 & 23.1 & 38.6  & 21.7 & 33.4 & 45.0  \\
&mr & 26.5 & 41.1 & 58.6 & 11.6 & 33.6 & 51.2  & 26.9 & 40.2 & 56.0 \\
&es & 19.6 & 31.8 & 48.8 & 09.0 & 17.6 & 40.8  & 20.8 & 31.7 & 45.6 \\

\bottomrule
\end{tabularx}
}%
\caption{Cross-lingual in-domain slot labelling results using XLM-R ($F_1$ $\times$ 100)}\label{tab: token classification  cross-lingual slot labelling}
\end{table*}

\begin{table*}[!htp]\centering
\scriptsize
\begin{tabularx}{\linewidth}{l YYY YYY YYY}
\toprule
  {} & \multicolumn{3}{c}{\bf \textsc{banking}} & \multicolumn{3}{c}{\bf \textsc{hotels}} & \multicolumn{3}{c}{\bf \textsc{all}} \\
  \cmidrule(lr){2-4} \cmidrule(lr){5-7} \cmidrule(lr){8-10}
\textbf{Model} & \textbf{20-Fold}  & \textbf{10-Fold} & \textbf{Large} & \textbf{20-Fold}  & \textbf{10-Fold} & \textbf{Large} & \textbf{20-Fold}  & \textbf{10-Fold} & \textbf{Large} \\ \midrule
\rowcolor{Gray} {} & \multicolumn{9}{c}{\bf \textsc{English}} \\ \midrule
mDeBERTa & 59.7 & 66.5 & 76.0 & 61.6 & 67.3 & 77.3 & 63.0 & 68.5 & 78.0 \\ \midrule
\rowcolor{Gray} {} & \multicolumn{9}{c}{\bf \textsc{amharic}} \\ \midrule
mDeBERTa & 03.2  & 03.8  & 05.0  & 05.8  & 06.8  & 09.9  & 03.5  & 03.9  &  05.4       \\ \midrule
\rowcolor{Gray} {} & \multicolumn{9}{c}{\bf \textsc{marathi}} \\ \midrule
mDeBERTa & 03.2  & 03.7  & 04.9  & 05.5  & 06.3  & 08.9  & 03.3  & 03.8  &   05.1     \\ \midrule
\rowcolor{Gray} {} & \multicolumn{9}{c}{\bf \textsc{spanish}} \\ \midrule
mDeBERTa & 46.0 & 52.4 & 60.1 & 52.6 & 58.7 & 68.8 & 52.5 & 57.9 &     65.7   \\ \midrule
\rowcolor{Gray} {} & \multicolumn{9}{c}{\bf \textsc{turkish}} \\ \midrule
mDeBERTa & 59.4 & 67.1 & 84.1 & 58.7 & 65.4 & 77.7 & 62.4 & 68.1 &   79.1  \\
\bottomrule
\end{tabularx}
\caption{In-domain in-language slot labelling with QA models ($F_1$ $\times$ 100)}\label{tab: in-domain in-language slot}
\end{table*}

\begin{table*}[!t]
\def\arraystretch{0.96}
\centering
{\scriptsize
\begin{tabularx}{\linewidth}{ll YYY YYY YYY}
\toprule
  {} & {} & \multicolumn{3}{c}{\bf \textsc{banking}} & \multicolumn{3}{c}{\bf \textsc{hotels}} & \multicolumn{3}{c}{\bf \textsc{all}}\\
  \cmidrule(lr){3-5} \cmidrule(lr){6-8} \cmidrule(lr){9-11} 
\textbf{SRC} & \textbf{TGT} & \textbf{20-Fold}  & \textbf{10-Fold} & \textbf{Large} & \textbf{20-Fold}  & \textbf{10-Fold} & \textbf{Large} & \textbf{20-Fold}  & \textbf{10-Fold} & \textbf{Large}\\
\cmidrule(lr){3-11}
\cmidrule(lr){1-11}

\multirow{5}{*}{en} & en & 59.7 & 66.5 & 76.0 & 61.6 & 67.3 & 77.3 & 63.0               & 68.5 & 78.0 \\
                    & am & 44.2 & 47.9 & 53.8 & 47.6 & 49.7 & 55.8 & 46.8               & 48.4 & 52.9 \\
                    & mr & 48.7 & 54.6 & 62.1 & 42.0 & 46.5 & 54.4 & 46.9               & 50.9 & 57.3 \\
                    & es & 42.9 & 48.0 & 55.1 & 49.0 & 54.1 & 60.7 & 47.7               & 51.7 & 58.1 \\
                    & tr & 52.3 & 58.4 & 69.0 & 50.9 & 55.8 & 63.0 & 53.6               & 58.4 & 65.2 \\
 \cmidrule(lr){1-11}
\multirow{5}{*}{am} & en & 2.5  & 2.8  & 3.6  & 4.1  & 4.3  & 6.1  & 2.4                & 2.6  & 3.4  \\
                    & am & 3.2  & 3.8  & 5.0  & 5.8  & 6.8  & 9.9  & 3.5                & 3.9  & 5.4  \\
                    & mr & 2.5  & 2.9  & 3.6  & 2.6  & 3.1  & 5.0  & 2.2                & 2.4  & 3.7  \\
                    & es & 1.8  & 2.0  & 2.7  & 3.1  & 3.2  & 4.9  & 1.8                & 2.0  & 2.5  \\
                    & tr & 2.9  & 3.4  & 4.4  & 3.6  & 4.1  & 5.8  & 2.6                & 2.9  & 4.3  \\
  \cmidrule(lr){1-11}
\multirow{5}{*}{mr} & en & 2.7  & 2.9  & 3.5  & 5.1  & 5.6  & 7.2  & 2.6                & 2.9  & 3.6  \\
                    & am & 2.6  & 2.9  & 3.8  & 4.2  & 4.9  & 6.3  & 2.6                & 2.9  & 3.7  \\
                    & mr & 3.2  & 3.7  & 4.9  & 5.5  & 6.3  & 8.9  & 3.3                & 3.8  & 5.1  \\
                    & es & 1.8  & 2.0  & 2.5  & 3.8  & 4.1  & 5.6  & 1.9                & 2.1  & 2.6  \\
                    & tr & 2.9  & 3.3  & 4.4  & 4.4  & 5.1  & 7.5  & 2.8                & 3.2  & 4.5  \\

\cmidrule(lr){1-11}

\multirow{5}{*}{es} & en & 54.5 & 61.7 & 69.9 & 54.8 & 59.3 & 66.9 & 55.5               & 60.2 & 68.0 \\
                    & am & 43.2 & 50.2 & 54.2 & 47.3 & 51.1 & 52.6 & 44.2               & 48.2 & 52.9 \\
                    & mr & 44.4 & 53.4 & 58.6 & 38.8 & 42.9 & 48.3 & 42.4               & 46.0 & 50.3 \\
                    & es & 46.0 & 52.4 & 60.1 & 52.6 & 58.7 & 68.8 & 52.5               & 57.9 & 65.7 \\
                    & tr & 50.0 & 58.0 & 66.0 & 48.4 & 53.2 & 59.0 & 50.6               & 54.5 & 62.1 \\
 \cmidrule(lr){1-11}

\multirow{5}{*}{tr} & en & 49.3 & 54.5 & 64.0 & 52.8 & 57.1 & 61.9 & 52.6               & 55.5 & 58.8 \\
                    & am & 44.4 & 51.7 & 56.5 & 45.8 & 49.8 & 55.1 & 45.5               & 49.9 & 51.8 \\
                    & mr & 47.9 & 55.9 & 68.0 & 38.5 & 43.1 & 51.5 & 45.2               & 50.5 & 58.9 \\
                    & es & 37.0 & 40.8 & 49.5 & 43.2 & 46.7 & 52.2 & 40.4               & 43.8 & 48.6 \\
                    & tr & 59.4 & 67.1 & 84.1 & 58.7 & 65.4 & 77.7 & 62.4               & 68.1 & 79.1 \\

\bottomrule
\end{tabularx}
}%
\caption{Cross-lingual in-domain slot labelling results using XLM-R ($F_1$ $\times$ 100)}\label{tab: qa based  cross-lingual slot labelling}
\end{table*}
\end{document}